\documentclass[aos,preprint]{imsart}

\RequirePackage[OT1]{fontenc}
\RequirePackage{amsthm,amsmath}
\RequirePackage[numbers]{natbib}
\RequirePackage[colorlinks,citecolor=blue,urlcolor=blue]{hyperref}
\usepackage{hhline,amssymb,epsfig,array,color,titlesec,lipsum,enumitem,booktabs,subcaption,lscape,float,stmaryrd}
\usepackage{xcolor}
\colorlet{darkblue}{blue!70!}

\arxiv{arXiv:0000.0000}

\startlocaldefs
\numberwithin{equation}{section}
\theoremstyle{plain}

\newtheorem{lemma}{Lemma}[section]
\newtheorem{theorem}{Theorem}[section]
\newtheorem{corollary}{Corollary}[section]
\newtheorem{assumption}{Assumption}[section]

\newtheorem{proposition}{Proposition}[section]

\endlocaldefs

\newcommand{\be}{\begin{eqnarray}}
\newcommand{\ee}{\end{eqnarray}}
\newcommand{\ba}{\begin{eqnarray*}}
\newcommand{\ea}{\end{eqnarray*}}

\newcommand{\argmin}{\arg\!\min}

\newcommand{\bx}{{\bf x}}

\newcommand {\bfdelta} {\mbox{\boldmath $\delta$}}

\newcommand {\bfTheta} {\mbox{\boldmath $\Theta$}}

\begin{document}

\begin{frontmatter}
\title{MULTILAYER TENSOR FACTORIZATION WITH APPLICATIONS TO RECOMMENDER SYSTEMS\thanksref{T1}}
\runtitle{MULTILAYER TENSOR FACTORIZATION}
\thankstext{T1}{Research is supported in part by National Science Foundation Grants DMS-1415500, 
DMS-1712564, DMS-1712564, DMS-1415308, DMS-1308227 and DMS-1415482, and National Institute on Drug Abuse 
Grant R01 DA016750.}

\begin{aug}
\author{\fnms{Xuan} \snm{Bi}\thanksref{m1}\ead[label=e1]{xuan.bi@yale.edu}},
\author{\fnms{Annie} \snm{Qu}\thanksref{m2}\ead[label=e2]{anniequ@illinois.edu}}
\and
\author{\fnms{Xiaotong} \snm{Shen}\thanksref{m3}\ead[label=e3]{xshen@umn.edu}}

\runauthor{X. BI, A. QU and X. SHEN}

\affiliation{Yale University\thanksmark{m1}, University of Illinois at Urbana-Champaign\thanksmark{m2} and University of Minnesota\thanksmark{m3}}

\address{Xuan Bi\\
Department of Biostatistics\\
Yale University\\
New Haven, Connecticut 06520\\
USA\\
\printead{e1}}

\address{Annie Qu\\
Department of Statistics\\
University of Illinois at Urbana-Champaign\\
Champaign, Illinois 61820\\
USA\\
\printead{e2}}

\address{Xiaotong Shen\\
School of Statistics\\
University of Minnesota\\
Minneapolis, Minnesota 55455\\
USA\\
\printead{e3}}
\end{aug}

\begin{abstract}
Recommender systems have been widely adopted by electronic commerce and entertainment industries 
for individualized prediction and recommendation, which benefit consumers and improve business intelligence. 
In this article, we propose an innovative method, {\color{black} namely the recommendation engine of multilayers (REM),} for tensor recommender systems.
The proposed method utilizes the structure of a tensor response to integrate information from multiple modes,
and creates an additional layer of nested latent factors to accommodate between-subjects dependency. One major advantage is that the proposed method is able to address the ``cold-start'' issue in
the absence of information from new customers, new products or new contexts. Specifically,
it provides more effective recommendations through sub-group information. 
To achieve scalable 
computation, we develop a new algorithm for the proposed method, which incorporates a
maximum block improvement strategy into the cyclic blockwise-coordinate-descent algorithm. 
{\color{black} In theory, we investigate both algorithmic properties for global and local convergence, along with the asymptotic consistency of estimated parameters.}
Finally,
the proposed method is applied in simulations and IRI marketing data with 116 
million observations of product sales. Numerical studies demonstrate that the proposed method outperforms existing competitors in the literature.
\end{abstract}



\end{frontmatter}

\section{Introduction}

Recommender systems have become very important in daily life due to high demand from the entertainment industry and business marketing which produce large amounts of data. In addition, applications of recommender systems have been greatly facilitated by the advancement of statistical and machine learning techniques. These applications include personalized marketing for internet users, merchandise recommendation for retail stores, and even individualized gene therapies. Each application involves collecting a wide variety of information, and successful exploitation of such rich information leads to more accurate recommendations. However, this also imposes unprecedented challenges to traditional methods due to the large size and complex structure of data. Therefore, more general and integrative recommender systems are urgently needed.

The tensor, also called multidimensional array, is well-recognized as a powerful tool to represent complex and unstructured data \citep{yuan2015tensor}. It is applied in many areas such as signal processing, neuroimaging, and psychometrics \citep[e.g.,][]{miranda2015tprm,li2016sparse,li2016parsimonious}. In recommender systems, the tensor shows its flexibility to accommodate contextual information, and is also regarded as one of the most effective tools for developing context-aware recommender systems \citep[CARS;][]{	adomavicius2005toward,adomavicius2011context}. \textcolor{black}{In addition to user and item information from traditional recommender systems, tensor-based recommender systems also take the effect of contextual variables into account, such as time, location, users' companions, stores' promotion strategies, other relevant variables, or any combinations thereof. Hence, CARS are capable of utilizing more information and provide more accurate recommendations \citep{verbert2012context,bobadilla2013recommender,shi2014collaborative}.}

Nevertheless, applying the tensor effectively to CARS remains a challenging problem. In matrix recommender systems, the singular value decomposition (SVD) method provides the best low-rank approximation, and is known to be arguably the most effective single procedure \citep{feuerverger2012statistical,nguyen2013content}. \textcolor{black}{In contrast, 
the SVD for tensor has more than one definition, and neither of the tensor decompositions inherits all of the desirable properties of a matrix SVD.}
This imposes a great challenge to generalize matrix decomposition to the tensor framework \citep{yuan2016incoherent}.

One common approach to utilize the tensor structure is to apply latent factor models, where each user, item or context is assigned an individual latent factor to represent their characteristics quantitatively. \textcolor{black}{Existing methods include, but are not limited to, the factor model with temporal dynamics \citep{koren2010collaborative} and the Bayesian probabilistic tensor factorization method \citep{xiong2010temporal}, both of which treat time as a contextual variable. Furthermore, the factorization machine \citep{rendle2011fast,rendle2012factorization,nguyen2014gaussian} models interactions of all possible pairs of variables, while the multiverse recommendation method applies Tucker decomposition \citep{karatzoglou2010multiverse}. Other existing methods for CARS include contextual pre- or post-filtering \citep{palmisano2008using,lombardi2009context}.}

However, several key issues have not been solved completely. The first is the ``cold-start'' problem, where available information is not sufficient to provide valid predictions for new users, items or contexts (in the rest of this paper, we use ``subject'' to denote a user, an item or a context in general). For instance, in latent factor modeling, a latent factor is not estimable if a subject is not available in the training set. The subject's utilities can only be predicted through the average information from other subjects, which may lead to low prediction accuracy. Several solutions have been proposed for the traditional matrix recommendation techniques, for example, imputing pseudo ratings \citep{goldberg2001eigentaste}, supplementing artificial users and items \citep{park2006naive}, incorporating content-boosted information \citep{forbes2011content,nguyen2013content}, or utilizing group information for new subjects \citep{bi2016group}. Nevertheless, the ``cold-start'' problem under CARS is quite challenging and has not been well-investigated. \textcolor{black}{One reason is that in addition to new users and new items, information on new contexts could be insufficiently collected as well.} For example, viewers may have different movie-watching experiences with new friends or at a new theater \citep{colombo2015recommetz}, and stores' sales volumes may vary under a new promotion strategy.

Another critical issue involves solving the higher-order tensors (beyond the third-order) problem. Higher-order tensors are very useful, because we might be interested in integrating more than one contextual variable to fully utilize their subject-specific information. For example, when recommending a new destination to travelers, one has to consider several factors, e.g., the timing and the cost of the trip, and the crowdedness of the destination, in addition to travelers' travel interests. While some existing methods provide a general methodology on high-order tensors \citep[e.g.,][]{karatzoglou2010multiverse}, the implementation of higher-order tensors could be challenging. One obstacle is the high computational cost. Another obstacle is that higher-order tensors could result in higher missing rates due to fewer observations at each combination of contextual variables; and this could lead to non-convergence in computing for some existing methods. One possible solution is to choose only one special contextual variable in the tensor, and treat the rest of the contextual variables as linear covariates \citep[e.g.,][]{koren2010collaborative,xiong2010temporal}. However, this may lead to loss of information on subject-specific or group-wise interaction.

In this paper, we propose a novel tensor factorization method, {\color{black} namely the recommendation engine of multilayers (REM)}.
Specifically, we assume a tensor structure where each mode corresponds to \textit{user}, \textit{item} or a \textit{contextual variable}, and each element of the tensor represents a utility, such as a rating or sales volume. The novelty of our method is that we add another layer which categorizes users, items and contexts into the same subgroups if they share similar characteristics. We quantify subgroup effects as random effects \citep{laird1982random,wang2012conditional} through nested-factors modeling, in addition to latent factors in the tensor factorization. Theoretically, we demonstrate the algorithmic properties of the proposed method, {\color{black}which converges to a stationary point from an arbitrary initial point,} and local convergence to a local minimum with a linear convergence rate. The estimated parameter achieves asymptotic consistency under the $L_2$-loss function and other more general circumstances. 

The proposed tensor factorization method has two significant advantages. First, it solves the ``cold-start'' problem effectively. For a new subject, even if its latent factors are not available, the nested factors from the corresponding subgroups can provide a group-specific estimate, which is more accurate than the average of the available observations. This finding is also supported by simulation studies where the proposed method is more effective than competing tensor factorization methods when the proportion of new subjects is high. 

Second, the proposed method is able to accommodate high-order tensors. The difficulty of applying high-order tensors is solved by the proposed nested factors which utilize group-wise information and hence are more robust to a higher missing rate. In addition, we propose a new algorithm that 
incorporates the maximum block improvement \citep{chen2012maximum} into the cyclic blockwise-coordinate-descent algorithm. This avoids the direct operation of large-scale tensors, and makes the estimation of high-order tensors feasible. Furthermore, a parallel computing strategy is implemented to calculate latent factors and nested factors for each subject and subgroup, respectively, which provides scalable and efficient computation.

The rest of this paper is organized as follows. Section \ref{Preparations} provides the background of the tensor factorization and the framework for the context-aware recommender systems. Section \ref{A Multilayer Method} introduces the proposed method and algorithm. Theoretical properties are derived in Section \ref{Theoretical Properties}. Section \ref{Simulation Studies} presents simulation studies to validate the performance of the proposed method. In Section \ref{realdata}, we apply the proposed method to IRI marketing data. Section \ref{Discussion} concludes with a discussion.

\section{Preparations} \label{Preparations}
\subsection{Notation and Tensor Background}
In this subsection, we provide the background of the tensor, and introduce the notation for the proposed method. 

We define a $d$-th order tensor $\mathbf{Y} \in \mathbb{R}^{n_1 \times \cdots \times n_d}$ as a $d$-dimensional array, where each order is also called a mode. 
In the rest of this article, bold capital letters denote tensors, capital letters denote matrices, bold small letters denote vectors, and small letters denote scalars.

\textcolor{black}{In contrast to the singular value decomposition for a matrix, 
the rank and the bases of a tensor cannot be obtained simultaneously. Two different decompositions are commonly adopted. One is the 
high-order singular value decomposition, which decomposes a tensor into a 
$d$-th order core tensor associated with $d$ orthonormal matrices. This decomposition provides a basis at each mode. However, the core tensor is usually 
non-diagonal, and the tensor rank is not estimable through high-order singular value decomposition. An alternative choice of decomposition is the Canonical Polyadic Decomposition (CPD), where a tensor is represented as a sum of $r$ rank-1 tensors.} That is:
\be \label{CP}
\mathbf{Y} \approx \sum_{j=1}^r \mathbf p^1_{\cdot j} \circ \mathbf p^2_{\cdot j} \circ \cdots \circ \mathbf p^d_{\cdot j},
\ee
where $\circ$ represents the vector outer product, and $\mathbf p_{\cdot j}^k$, $j=1,\ldots,r$, are $n_k$-dimensional vectors corresponding to the $k$-th mode.
Here $r$ is called the rank of $\mathbf{Y}$ if the number of terms $r$ is minimal. 
Equivalently, for each element of $\mathbf{Y}$, we have:
\be \label{CPelement}
y_{i_1i_2\cdots i_d} \approx \sum_{j=1}^r p^1_{i_1j} p^2_{i_2j} \cdots p^d_{i_dj}.
\ee

\textcolor{black}{In the context of recommender systems, 
the tensor decomposition technique is geared towards the 
interpretation of intrinsic data variation. Therefore, we employ
the CPD because the rank of a tensor corresponding to the number of latent factors is more important than the 
orthonormality.}
For other properties of a tensor, see \cite{kolda2009tensor} for an extensive review.



\subsection{Context-Aware Recommender Systems} \label{CARS22}
In this subsection, we briefly review the tensor application to recommender systems, namely, context-aware recommender systems (CARS). See also \cite{adomavicius2011context} for a comprehensive review of CARS.


\textcolor{black}{Consider a $d$-th order tensor ($d\ge3$) with the first two modes corresponding to \textit{user} and \textit{item} and the other $(d-2)$ modes corresponding to \textit{contextual variables}.} Let $n_k$ $(k=1,\ldots,d)$ be the number of subjects for the $k$-th mode, that is, $n_1$ is the number of users, $n_2$ is the number of items, and $n_3,\ldots,n_d$ are the number of contexts for the $(d-2)$ contextual variables, respectively. For the CPD representation in (\ref{CP}), let $P^k=(\mathbf p^k_{\cdot 1},\ldots,\mathbf p^k_{\cdot r})_{n_k \times r}$, where each row of $P^1$ or $P^2$ represents the $r$-dimensional latent factors for each user or item, and rows of $P^3,\ldots,P^d$ are the latent factors for the contextual variables, respectively. In the rest of this article, we use $\mathbf p^k_{i_k}$ to represent the $i_k$-th row of $P^k$, in contrast to $\mathbf p_{\cdot j}^k$ being the $j$-th column of $P^k$. Each element of $\mathbf{Y}$ is defined as a utility, for example, a rating, a purchase or a sales volume, and is estimated via (\ref{CPelement}) in CPD.
Here the orthonormality for the latent factors is not required, and each utility $y_{i_1i_2\cdots i_d}$ comprises user-, item- and context-specific information. In addition, the CPD requires estimation of only $\sum_{k=1}^d n_kr$ parameters, which essentially performs a tensor dimension reduction procedure.

In many applications, we may have a non-negative tensor $\mathbf Y$. Methods for non-negative CPD are proposed to address this issue \citep[e.g.;][]{paatero1999multilinear,welling2001positive,chi2012tensors,aswani2016low}, as such decompositions make results meaningful and interpretable. In recommender system problems, however, the direct interpretation of latent factors is less critical, as the relative scale or ranking is important for recommendation. Most importantly, improving prediction accuracy is the ultimate goal of a good recommender system for users. Therefore, non-negativity is not required here. Technically, we can always standardize $\mathbf Y$ prior to analysis. In addition, non-negative CPDs usually entail non-negative latent factors, which might restrict the parameter space to the non-negative orthant. In the proposed framework, an unbounded parameter space allows more flexibility for extensive search, which may lead to a more satisfactory convergence result.

The most common loss function is the $L_2$-loss, which is computationally efficient. Theoretical properties of other loss functions are also considered \citep[e.g.,][]{srebro2005generalization,zhu2016personalized}. Let $\Omega=\{(i_1,i_2,\ldots,i_d): y_{i_1i_2\cdots i_d} \mbox{ is observed}\}$ be a set of indices corresponding to observed utilities, and $|\Omega|$ represent the sample size. Since the number of available observations for each subject might be smaller than the number of latent factors $r$, we adopt a penalty function using regularization. For example, when the $L_2$-penalty is applied, we have:
\ba \label{CPcriterion}
L(P^1,\ldots,P^d|\mathbf Y)=\sum_{(i_1,\cdots,i_d) \in \Omega}(y_{i_1\cdots i_d} - \hat{y}_{i_1\cdots i_d})^2 +\lambda \sum_{k=1}^d\|P^k\|_F^2,
\ea
where $\hat{y}_{i_1\cdots i_d}=\sum_{j=1}^r p^1_{i_1j} \cdots p^d_{i_dj}$ is the estimated utility provided by (\ref{CPelement}), and $\|\cdot\|_F$ represents the Frobenius norm. Other regularization methods include, but are not limited to, 
the trend filtering penalty when $d=3$ \citep{xiong2010temporal},
and the $L_0$- and $L_1$-penalty for sparse low-rank pursuit when $d=2$ 
\citep{zhu2016personalized}.

Algorithms for implementing the CPD include the cyclic coordinate descent algorithm, and the stochastic gradient descent method. Alternatively, \cite{chen2012maximum} propose a maximum block improvement algorithm that only updates the block with the maximum improvement in each iteration instead of updating each block cyclically. This strategy guarantees convergence to a stationary point, and ensures a fast convergence rate in many circumstances.


\section{A Multilayer Method} \label{A Multilayer Method}
\subsection{General Methodology} \label{General Methodology}
In this section, we develop the methodology for the proposed REM method. 
Specifically, we adopt the idea of nested factors from the design of experiments to capture between-subject dependency under the CPD framework (\ref{CP}).

We assume that subjects can be categorized into subgroups, where subjects within the same subgroup share similar characteristics and are dependent on each other. For subgrouping, we can incorporate prior information such as users' demographic information, item categories and functionality, and contextual similarity. If this kind of information is not available, one can utilize the missing pattern of the tensor data, or the number of records from each user and on each item \citep{salakhutdinov2007restricted}. As shown in \cite{feuerverger2012statistical}, implicit information from the number of records may reflect subjects' behavior that is not available elsewhere, and can help improve recommendation accuracy. In more general situations, clustering methods such as the $k$-means can be applied to determine the subgroups. See \cite{wang2010consistent} and \cite{fang2012selection} on robust approaches to select the number of subgroups.

Suppose the subgroup labels are given, then we formulate each utility as follows:
\be \label{GSM}
\hat{y}_{i_1i_2\cdots i_d} = \sum_{j=1}^r (p^1_{i_1j}+q^1_{i_1j}) (p^2_{i_2j}+q^2_{i_2j}) \cdots (p^d_{i_dj}+q^d_{i_dj}),
\ee
where $p^k_{i_kj}$ is the $j$-th latent factor for the $i_k$-th subject from the $k$-th mode, and $q^k_{i_kj}$ is the corresponding nested factor, $j=1,\ldots,r$, $i_k=1,\ldots,n_k$, and $k=1,\ldots,d$. 
{\color{black}
We define the $n_k \times r$-dimensional matrix $Q^k$ similar to $P^k$ as in Section \ref{CARS22}. Notice that we have $\mathbf q^k_{i_k}=\mathbf q^k_{i'_k}$ if subjects $i_k$ and $i'_k$ are from the same subgroup. We assume that the number of subgroups for the $k$-th mode is $m_k$, which corresponds to $m_k$ unique values for $q^k_{i_kj}$. 
We use $\mathbf q_{(u_k)}^k$ occasionally to denote the nested factor associated with the subgroup $u_k$, $u_k=1,\ldots,m_k$.  
}

Let $P=\left((P^1)',\ldots,(P^d)'\right)'$ and $Q=\left((Q^1)',\ldots,(Q^d)'\right)'$ represent all parameters of interest. We define $L(P,Q|\mathbf Y)=L(P^1,\ldots,P^d,Q^1,\ldots,Q^d|\mathbf Y)$ as the overall criterion function:
\be \label{pracloss}
L(P,Q|\mathbf Y)=
\sum_{(i_1,\cdots,i_d) \in \Omega} (y_{i_1\cdots i_d} - \hat{y}_{i_1\cdots i_d})^2 +\lambda (\|P\|_F^2+\|Q\|_F^2),
\ee
where $\hat{y}_{i_1\cdots i_d}$ is represented by (\ref{GSM}) and $\lambda$ is the penalization coefficient. Here we adopt the most commonly 
used $L_2$-loss and $L_2$-penalty for efficient computation, although other types of loss and penalty functions are also applicable. For example, one may consider the hinge loss or the $\psi$-loss \citep{shen2003psi} 
for classification, and the absolute loss or Huber loss to achieve robust estimation. Meanwhile, if prior knowledge of the latent and nested factors is available, for example regarding sparsity or smoothness, then appropriate regularization methods can be applied.

\subsection{Parameter Training}
{\color{black}
In the following, we discuss how the model parameters are estimated.
Mainly we are interested in finding a solution $(\hat{P},\hat{Q})$ aiming at minimizing $L(P,Q|\mathbf Y)$.
}
Let $\Omega^k_{i_k}=\{(i_1,\ldots,i_k,\ldots,i_d): \mbox{ }y_{i_1\cdots i_k\cdots i_d} \mbox{ is observed}\}$ be the set of indices where the $k$-th mode index equals $i_k$ and the corresponding utilities are observed; \textcolor{black}{namely, $|\Omega^k_{i_k}|$ denotes the number of observations for subject $i_k$.} Let $\mathcal{I}^k_{(u_k)}$ be the set of subjects in the subgroup $u_k$, $u_k=1,\ldots,m_k$. We assume that $|\mathcal{I}^k_{(u_k)}| \ge 2$ for each $u_k$.

For each mode of the tensor, the partial derivatives of $L(\cdot|\mathbf Y)$ have explicit forms with respect to the latent factors or the nested factors, which makes it feasible to apply the blockwise coordinate descent approach.
That is,
\be \label{estp}
 \hat{\mathbf p}^k_{i_k} = \mathop{\argmin}_{\mathbf p^k_{i_k}} \sum_{\Omega^k_{i_k}} (y_{i_1\cdots i_d} - \hat{y}_{i_1\cdots i_d})^2 +\lambda \|\mathbf p^k_{i_k}\|_2^2,
\ee
for $i_k=1,\ldots,n_k,$ and
\be \label{estq}
 \hat{\mathbf q}^k_{(u_k)} = \mathop{\argmin}_{\mathbf q^k_{(u_k)}} \sum_{i_k \in \mathcal{I}^k_{u_k}}\sum_{\Omega^k_{i_k}} (y_{i_1\cdots i_d} - \hat{y}_{i_1\cdots i_d})^2 +\lambda \|\mathbf q^k_{(u_k)}\|_2^2,
\ee
for $u_k=1,\ldots,m_k$, and $k=1,\ldots,d$.

The estimation procedure of $\hat{\mathbf p}^k_{i_k}$ in (\ref{estp}) is a ridge regression, and does not require knowing $\hat{\mathbf p}^k_{i'_k}$ for $i'_k \ne i_k$. Thus, parallel computation is applicable to calculate $\hat{\mathbf p}^k_1,\ldots,\hat{\mathbf p}^k_{n_k}$ efficiently. This strategy is also applicable to obtaining the $\hat{\mathbf q}^k_{(u_k)}$'s in (\ref{estq}). Therefore, the minimization of $L(P,Q|\mathbf Y)$ can be done cyclically through estimating $P$ and $Q$.

Notice that $\Omega =\cup_{i_k=1}^{n_k}\Omega^k_{i_k}$, and it is possible that $\Omega_{i_k}^k$ is empty for certain $i_k$'s; that is, there is no observation on subject $i_k$, as in the case of the ``cold-start'' problem. \textcolor{black}{Under this circumstance, the latent factor of $i_k$ is not estimable, and is assigned as $\mathbf p^k_{i_k}=\mathbf 0$. The predicted values calculated by existing methods may degenerate to the grand mean or subjects' main effects. In contrast, the proposed method utilizes the nested factor $\mathbf q^k_{(u_k)}$, which borrows information from members of the same subgroup. Thus, even for a new subject, the predicted values calculated by
(\ref{GSM})
retain information from other modes, and hence achieve better prediction accuracy.}

\subsection{Algorithm}
In contrast to matrix factorization, \textcolor{black}{a tensor decomposition usually entails high computational cost, and hence many algorithms feasible for a matrix may not be scalable for tensor decomposition. For example, it is nearly impossible to embed the back-fitting algorithm into the maximum block improvement (MBI) for tensor data}. First, the number of parameters and sample size for a tensor decomposition could be much greater than its matrix counterpart, which makes the computation of the MBI more intensive. Second, since the number of blocks increases significantly, the MBI algorithm may never update certain blocks due to small improvements along these directions, which leads to the estimated values corresponding to these blocks remaining the same as the initial values. To solve these problems, we propose a two-step algorithm, which estimates the latent-factors matrix $\hat{P}$ and the nested-factors matrix $\hat{Q}$ iteratively. Within the estimation of each matrix, we apply the MBI algorithm to find the optimal block direction with the largest improvement of estimations.

Specifically, we propose the following algorithm aiming at minimizing (\ref{pracloss}). Let $(P_s,Q_s)$ denote the estimated $(P,Q)$ at the $s$-th iteration, then the improvement of estimations for updating the $k$-th mode is defined as
\be \label{impI}
I^k_s=1-\frac{L(P^1_{s-1},\ldots,P^{k-1}_{s-1},P^{k*},P^{k+1}_{s-1},\ldots,P^{d}_{s-1},Q_{s-1}|\mathbf Y)}{L(P_{s-1},Q_{s-1}|\mathbf Y)},
\ee
and
\be \label{impJ}
J^k_s=1-\frac{L(P_{s-1},Q^{1}_{s-1},\ldots,Q^{k-1}_{s-1},Q^{k*},Q^{k+1}_{s-1},\ldots,Q^{d}_{s-1}|\mathbf Y)}{L(P_{s-1},Q_{s-1}|\mathbf Y)},
\ee
where $P^{k*}$ and $Q^{k*}$ are the attempted updates for the $k$-th mode, $k=1,\ldots,d$.

\noindent $\overline{\mbox{\underline{\makebox[\textwidth]{\textbf{Algorithm 1:} A Two-Step Algorithm with Parallel Computing}}}}$
\begin{enumerate}[topsep=-3mm, itemsep=-0mm]
\singlespacing
\item \textit{(Initialization)} Input all observed $y_{i_1 \cdots i_d}$'s, the rank $r$, the tuning parameter $\lambda$, initial value $(P_0,Q_0)$ and a stopping criterion $\varepsilon=10^{-4}$.

\item \textit{(Latent-factors update)} At the $s$-th iteration $(s\ge1)$, estimate $P_s$.

\begin{enumerate}[topsep=-0mm, itemsep=-0mm]
\item[(i)] For each $P^{k}$, solve (\ref{estp}) through parallel computing and obtain $P^{k*}=(\hat{\mathbf p}^k_{1},\ldots,\hat{\mathbf p}^k_{n_k})'$. 
Calculate $I^k_s$ through (\ref{impI}).

\item[(ii)] Assign $P^{k_0}_s \leftarrow P^{k_0*}$, if $I^{k_0}_s=\max \{I^1_s,\ldots,I^d_s\}$.
\end{enumerate}

\item \textit{(Nested-factors update)} At the $s$-th iteration $(s\ge1)$, estimate $Q_s$.
\begin{enumerate}[topsep=-0mm, itemsep=-0mm]

\item[(i)] For each $Q^{k}$, solve (\ref{estq}) through parallel computing and obtain $Q^{k*}=(\hat{\mathbf q}^k_{1},\ldots,\hat{\mathbf q}^k_{n_k})'$.
Calculate $J^k_s$ through (\ref{impJ}).

\item[(ii)] Assign $Q^{k_0}_s \leftarrow Q^{k_0*}$, if $J^{k_0}_s=\max \{J^1_s,\ldots,J^d_s\}$.
\end{enumerate}

\item \textit{(Stopping Criterion)} Stop if $$\max \{I^1_s,\ldots, I^d_s,J^1_s,\ldots,J^d_s\} < \varepsilon.$$
Set $(\hat{P}^1,\ldots,\hat{P}^d,\hat{Q}^1,\ldots,\hat{Q}^d) = (P^{1}_s,\ldots,P^{d}_s,Q^{1}_s,\ldots,Q^{d}_s)$. 

Otherwise set $s \leftarrow s+1$ and go to step 2.
\end{enumerate}
\noindent\makebox[\linewidth]{\rule{\textwidth}{0.4pt}}

One advantage of the proposed algorithm is that it requires small memory storage. Note that at each iteration, only one subject's information is required to estimate $\hat{\mathbf p}^k_{i_k}$ and one sub-group's information is required to estimate $\hat{\mathbf q}^k_{(u_k)}$. Furthermore, the computational complexity of the proposed algorithm is no greater than $2dn_{iter}c_{ridge}$, where $n_{iter}$ is the number of iterations and $c_{ridge}$ is the complexity of the ridge regression. In addition, since the MBI algorithm does not update blocks cyclically, it is able to discover and utilize ``shortcuts'' in optimization, which may significantly reduce the number of iterations.

\subsection{Implementation} \label{Implementation}

In this subsection, we address several implementation issues. In general, we split the data into a 50\% training set, a 25\% validation set and a 25\% testing set, randomly. The tuning parameter $\lambda$ is selected to minimize the root mean square error (RMSE) on the validation set, where the RMSE on a set $\Omega$ is defined as $\sqrt{\frac{1}{|\Omega|}\sum_{\Omega}(y_{i_1\cdots i_d}-\hat{y}_{i_1\cdots i_d})^2}$. 
{\color{black}
To improve prediction, we could specify $\lambda$ differently for each row of $P$ and $Q$. In our numerical study, we use $\lambda$ uniformly for the latent factors in $P$, but use $\lambda^k_{(u_k)}=\lambda/|\mathcal{I}_{(u_k)}^k|$ for each subgroup $u_k$ to penalize each parameter equally. 
}
Furthermore, we need to choose the number of latent factors $r$. In general, $r$ is no smaller than the theoretical rank of the tensor in order to represent subjects' characteristics sufficiently well. However, a large $r$ may lead to intensive computation and possible non-convergence of algorithms.

Moreover, it is important to determine which contextual variables should be integrated in the tensor. In Section \ref{hotensor}, we demonstrate that if the true tensor is of high-order, then assuming a low-order tensor structure may lead to a loss of information. That is, it is important to include key contextual variables in the tensor. On the other hand, applying a high-order tensor to formulate a low-order problem is unnecessary and may entail extra computational cost. In practice, we assume that the order of a tensor can be judged from prior knowledge. We acknowledge that determining the order of a tensor remains an open problem.



{\color{black}
{\color{black}
\section{Theoretical Properties} \label{Theoretical Properties}

This section develops theoretical properties for the proposed method. 
Our contributions are mainly on two aspects. One is on the convergence properties of the proposed algorithm, {\color{black}which converges to a stationary point from an arbitrary initial point,} and local convergence to a local minimum (or a global minimum, if one exists) with linear convergence rate. 
The other contribution is on the statistical properties. We prove the asymptotic consistency of the estimated parameter under $L_2$ loss and more general criterion functions.

One well-known critical issue is the discrepancy between the algorithmic and statistical properties \citep[e.g.;][]{zhu2016personalized,bi2016group}. In some existing works, the statistical framework may require the estimated parameter to be a global minimizer, which might not be existent or attainable. 

Our theoretical development, nevertheless, bridges this gap from two aspects. First and foremost, we relax the strict condition such that a global minimizer is no longer required to establish statistical properties, as long as the criterion function converges to its infimum asymptotically. Second, we provide technical solutions on finding possible global minima or satisfactory local minima, with additional computational cost. 

\subsection{Identifiability}

Identifiability is critical for tensor representations.
For recommender systems, although having identifiable latent factors does not improve prediction accuracy, it could still be important for algorithmic convergence which may lead to favorable statistical properties. Here we provide sufficient conditions to achieve identifiable latent factors prior to establishing theoretical properties.

In the proposed framework, unidentifiability is attributed to four aspects. The first three aspects are elementary indeterminacies of scaling, permutation and addition, whereas the 
last one is the so called non-uniqueness of the CPD with  more than one possible 
combination of rank-one tensors sum to $\mathbf{Y}$ after controlling for the three elementary indeterminacies \citep{kolda2009tensor}.
}

 Let $B=\left((B^1)',\ldots,(B^d)'\right)'$ where $B^k=P^k+Q^k$, $k=1,\ldots,d$. The 
scaling indeterminacy refers to non-uniqueness with respect to a scale change of 
each column vector of $B^k$. That is, for $d$ diagonal scaling matrices $\Gamma^k=\mbox{diag}(\gamma^k_1,\ldots,\gamma^k_r)$, $k=1,\ldots,d$, we have $\tilde{B} =\left((B^1 \Gamma^1)',\ldots,(B^d \Gamma^d)'\right)'$ such that $\prod_{k=1}^d \gamma^k_j=1$ for $j=1,\ldots,r$.
The permutation indeterminacy comes from an arbitrary $r \times r$ permutation matrix $\Pi$, 
such that $\tilde{B} =\big((B^1 \Pi)', \ldots,$ $(B^d \Pi)'\big)'$. In addition to scaling and 
permutation, the proposed tensor representation may suffer from the addition indeterminacy, that is,
for arbitrary $n_k \times r$ matrix $\Delta^k$, 
$\tilde{P}^k=P^k+\Delta^k$ and $\tilde{Q}^k=Q^k-\Delta^k$. 

 As a special case of $d=2$,  the elementary indeterminacy reduces to the non-singular 
transformation indeterminacy for a matrix. Specifically, for an $r \times r$ non-singular matrix $\Upsilon$, $\tilde{B}^1=B^1 \Upsilon$ and $\tilde{B}^2=B^2 (\Upsilon^{-1})'$. Nevertheless, 
this issue can always be solved by the singular value decomposition, which 
imposes orthonormality to column vectors. Therefore, we focus our attention on
higher-order tensors with $d \ge 3$, although some of the results continue to hold
for a matrix.

\begin{lemma} \label{invariant}
Predicted values given by (\ref{GSM}) are invariant with respect to scaling, 
permutation and addition indeterminacies.
\end{lemma}

The proof is straightforward by applying the aforementioned definition of scaling, permutation and addition indeterminacies to (\ref{GSM}), and is hence skipped. 
We introduce the concept of \emph{$k$-rank}, which is the Kruskal rank introduced
in \cite{kruskal1977three}. Specifically, for a matrix $A$, the $k$-rank of 
$A$ is 
$$\mathcal{K}_A=\max \{k:\mbox{ any $k$ columns of $A$ are linearly independent}\}.$$

\begin{proposition} \label{identifiability}
Suppose $\sum_{k=1}^d \mathcal{K}_{B^k} \ge 2r+(d-1)$. Minimizers  
of $L(P,Q|\mathbf{Y})$ in $P$ and $Q$  are unique up to permutation almost surely.
\end{proposition}

  The above condition $\sum_{k=1}^d \mathcal{K}_{B^k} \ge 2r+(d-1)$ is not strong. 
In numerical studies, factors in $(P^k+Q^k)$ are usually linearly independent, and hence we have $\mathcal{K}_{B^k}=r$. Then this
condition reduces to $r \ge 1+ 1/(d-2)$ for $d \ge 3$, which is achievable even for 
low-rank high-order tensors.

  As shown in Proposition \ref{identifiability}, the issue of scaling {\color{black}and addition indeterminacies}
is resolved almost surely through penalization imposed on $L(P,Q|\mathbf{Y})$. To
treat the permutation indeterminacy, we rearrange $r$ column vectors 
$(\mathbf{p}_{\cdot 1}^k+\mathbf{q}_{\cdot 1}^k, \mathbf{p}_{\cdot 2}^k+\mathbf{q}_{\cdot 2}^k, \ldots,\mathbf{p}_{\cdot r}^k+\mathbf{q}_{\cdot r}^k)$ for each mode such that 
$$\sum_{k=1}^d\|\mathbf{p}_{\cdot 1}^k+\mathbf{q}_{\cdot 1}^k\|_2^2 \ge \sum_{k=1}^d\|\mathbf{p}_{\cdot 2}^k+\mathbf{q}_{\cdot 2}^k\|_2^2 \ge \cdots \ge \sum_{k=1}^d\|\mathbf{p}_{\cdot r}^k+\mathbf{q}_{\cdot r}^k\|_2^2,$$
which is analogous to imposing a descending order of eigenvalues as in matrix decomposition. 
The rearrangement of column vectors can be implemented during or after the proposed algorithm, since it does not affect the estimation procedure.
We acknowledge that the above choice is arbitrary.
One may carry out other rearrangements. For example, 
\cite{zhou2013tensor} suggest imposing ordering based on the first element rather than the vector norm.
Alternatively, one may utilize prior knowledge if available. In the rest of Section \ref{Theoretical Properties}, we assume that parameter $(P,Q)$ is identifiable.



\subsection{Algorithmic Properties} \label{Algorithmic Properties}

\subsubsection{{\color{black}Convergence with an Arbitrary Initial Point}}

This subsection investigates the convergence property of the proposed algorithm in 
terms of $(P,Q)$ {\color{black}from an arbitrary initial point}. We first establish
the property in a compact parameter space and then generalize it to an unbounded open
parameter space with extra conditions.

Let $\mathcal{D} \subset \mathbb{R}^{2r\sum_{k=1}^d n_k}$ be the parameter space of $(P,Q)$. Then $(\tilde{P},\tilde{Q}) \in \mathcal{D}$ is called a \textit{blockwise local minimizer} of $L(\cdot|\mathbf Y)$ if
\[\tilde{P}^k=\mathop{\argmin}_{P^k}L(\tilde{P}^1,\ldots,\tilde{P}^{k-1},P^k,\tilde{P}^{k+1},\ldots,\tilde{P}^d,\tilde{Q}^1,\ldots,\tilde{Q}^d|\mathbf Y),\]
and
\[\tilde{Q}^k=\mathop{\argmin}_{Q^k}L(\tilde{P}^1,\ldots,\tilde{P}^d,\tilde{Q}^1,\ldots,\tilde{Q}^{k-1},Q^k,\tilde{Q}^{k+1},\ldots,\tilde{Q}^d|\mathbf Y),\]
which is equivalent to a local minimizer along each block direction. In the rest of Section \ref{Algorithmic Properties}, we consider the criterion function $L(\cdot|\mathbf Y)$ as defined in (\ref{pracloss}). 


\begin{lemma} \label{stationary}
Suppose $\mathcal{D}$ is 
compact, and the iterates obtained 
from Algorithm 1 have a cluster point $(\tilde{P},\tilde{Q})$. Then $(\tilde{P},\tilde{Q})$ is a blockwise local minimizer of the criterion function $L(\cdot|\mathbf Y)$.
\end{lemma}

In the following, we assume that $\mathcal{D}$ is open.
We consider $L(\cdot|\mathbf Y)$ at each of its block coordinates. Let 
$$L^k(P^k)=L(P^k|\mathbf Y,P^{(-k)},Q) \mbox{ and } L^{k+d}(Q^k)=L(Q^k|\mathbf Y,P,Q^{(-k)})$$ be $L(P,Q|\mathbf Y)$ given $(P^{(-k)},Q)$ and $(P,Q^{(-k)})$, respectively, where $X^{(-k)}=(X^1,\ldots,X^{k-1},X^{k+1},\ldots,X^d)$ for $X=P$ or $Q$, $k=1,\ldots,d$. 
Then Assumption \ref{homo} implies that the improvement on ($P^k$)'s or ($Q^k$)'s is not dominated by each other.

\begin{assumption}  \label{homo}
Let $\{(P_s,Q_s)\}_{s \ge 1}$ be a sequence of estimated parameters generated by Algorithm 1, where $s$ represents the $s$-th iteration. 
Then
$$O\left(\max_{k=1,\ldots,d} \|\nabla L^k(P^k_s) \|_F \right) \sim O\left(\max_{k=1,\ldots,d} \|\nabla L^{k+d}(Q^k_s) \|_F \right).$$

\end{assumption}

Furthermore, let $H(L^k)$ be the Hessian matrix of $L^k(\cdot)$, $k=1,\ldots,2d$.
We assume that $\|H(L^k)\|_2 \le \zeta^k$,
where the constant $\zeta^k>0$ is bounded above and may depend on all block coordinates of $L(\cdot|\mathbf{Y})$ except the $k$-th block.
The following proposition leads to {\color{black}the convergence to a blockwise local minimizer}.

\begin{proposition} \label{gc} 
Suppose $\mathcal{D}$ is open and Assumption \ref{homo} holds.
Let $\|H(L^k)\|_2 \le \zeta^k$ for $k=1,\ldots,2d$. Then the sequence $\{(P_s,Q_s)\}_{s \ge 1}$ obtained from Algorithm 1 converges to a blockwise local minimizer 
of the criterion function $L(P,Q|\mathbf Y)$.
\end{proposition}

Notice that, since the parameter space is open, a blockwise local minimizer satisfies
$\nabla L=\mathbf{0}$, and hence is a special case of a stationary point.

\subsubsection{Local Convergence}

In this subsection, we provide the local convergence property of Algorithm 1. 
Specifically, we follow \cite{li2015convergence} and show that Algorithm 1 converges to a local minimum at
the linear rate, provided that an initial value is 
sufficiently close to the local minimizer. Moreover, 
the same property applies to a global minimum if it exists. 



Let $(\tilde{P},\tilde{Q})$ be a local minimizer of $L(P,Q|\mathbf{Y})$, and $\tilde{H}=H\left(L(\tilde{P},\tilde{Q}|\mathbf{Y})\right)$ be the Hessian matrix at $(\tilde{P},\tilde{Q})$. We define the \textit{energy norm} based on $\tilde{H}$ as $\|(P,Q)\|_E=\left\langle \mbox{vec}(P,Q), \tilde{H}\mbox{vec}(P,Q) \right\rangle^{1/2}$.

\begin{proposition} \label{lc}
Suppose $\mathcal{D}$ is open, and let $(\tilde{P},\tilde{Q}) \in \mathcal{D}$ be a strict local minimizer of $L(P,Q|\mathbf{Y})$. For a small neighborhood $\mathcal{V}$ of $(\tilde{P},\tilde{Q})$, suppose $(P_{s_0},Q_{s_0}) \in \mathcal{V}$ for some $s_0 \ge 0$. Then a sequence $\{(P_s,Q_s)\}_{s \ge s_0} \subset \mathcal{V}$ obtained from Algorithm 1 exists, and converges to $(\tilde{P},\tilde{Q})$ at least linearly in the energy norm. That is, there exists $\mu \in [0,1)$, such that 
$$\|(P_{s+1},Q_{s+1})-(\tilde{P},\tilde{Q})\|_E \le \mu \|(P_{s},Q_{s})-(\tilde{P},\tilde{Q})\|_E.$$
\end{proposition}

Suppose the tolerance error of Algorithm 1 is $\epsilon$, and $s_0=0$. Then the number of iterations has an upper bound:
$$n_{iter} \sim O\left(\left\{\log \epsilon - \log \|(P_{0},Q_{0})-(\tilde{P},\tilde{Q})\|_E\right\} / \log \mu\right).$$

Several methods  are available to obtain a good initial value. One suggestion is to increase the sample size $|\Omega|$ \citep{bhojanapalli2015new,yuan2015tensor}. For an $n_1 \times \cdots \times n_d$-dimensional tensor, the sample size can potentially reach $|\Omega|=n_1\cdots n_d$, while the number of parameters is $r\sum_{k=1}^d n_k$. Since each step of the algorithm is a ridge regression, a larger sample size leads to more accurate estimation. However, sample size is not the only determinant. As discussed in \cite{de2008tensor} and \cite{kolda2009tensor}, there does not exist a best rank-$r$ approximation for high-order tensors in general. 
Alternatively, one could employ the branch-and-bound technique \citep{land1960automatic,clausen1999branch}, or utilize multiple random start points. 
These techniques could result in a satisfactory local minimum if computational capacity allows.

Since the parameter space is open, a global minimum is also a local minimum if it exists. Therefore, Proposition \ref{lc} applies if an initial value is in a small neighborhood of the global minimum.

\begin{corollary} \label{lc2}
Suppose $\mathcal{D}$ is open, and that a global minimizer $(\tilde{P},\tilde{Q})$ of $L(P,Q|\mathbf{Y})$ exists. Let $\tilde{H}$ be positive definite. For a small neighborhood $\mathcal{V}$ of $(\tilde{P},\tilde{Q})$, suppose $(P_{s_0},Q_{s_0}) \in \mathcal{V}$ for some $s_0 \ge 0$, then a sequence $\{(P_s,Q_s)\}_{s \ge s_0} \subset \mathcal{V}$ obtained from Algorithm 1 exists, and converges at least linearly to $(\tilde{P},\tilde{Q})$ in the energy norm. That is, there exists $\mu \in [0,1)$, such that 
$$\|(P_{s+1},Q_{s+1})-(\tilde{P},\tilde{Q})\|_E \le \mu \|(P_{s},Q_{s})-(\tilde{P},\tilde{Q})\|_E.$$
\end{corollary}

{\color{black}
The proof is a straightforward exercise of  Proposition \ref{lc} 
given that $\nabla L=\mathbf{0}$ and $\tilde{H}$ is positive.
Nevertheless, the existence of a global minimum is not guaranteed. For instance, 
as demonstrated in \cite{paatero2000construction} and \cite{de2008tensor},
a tensor might be approximated arbitrarily well by lower-rank tensors. \cite{de2008tensor} also demonstrate that the best rank-$r$ approximation problem could be ill-posed: for tensors with some rank $r$ in certain tensor spaces, there is a strictly positive probability that a global minimum cannot be obtained. For the recommender system framework, this issue is further complicated since a large proportion of tensor entries are missing. In the proposed setting, however, the criterion function $L(\cdot|\mathbf Y)$ is always bounded below by zero. Therefore, even if a global minimum does not exist, we can still minimize $L(\cdot|\mathbf Y)$ such that it is sufficiently close to its infimum.



\subsection{Asymptotic Properties}

In this subsection, we derive asymptotic properties for the proposed method. Specifically, we prove consistency of estimated parameters when the sample size goes to infinity. The result holds true under the $L_2$-loss, or more general loss functions with additional smoothness conditions.
As illustrated in the previous section, the global minimum of the criterion function may not exist, but the criterion function is bounded below by zero. We demonstrate that our asymptotic properties still hold even if the estimated parameter is not a global minimizer, as long as the criterion function converges to its infimum.

\subsubsection{Consistency under the $L_2$-Loss}

In this section, we focus on the asymptotic properties of the predicted values instead of the latent and nested factors $(P,Q)$, since prediction accuracy instead of parameter estimation is the primary concern. 
}

For this purpose, we re-define the parameter space. Suppose $\mathbf Y$ is a $d$-th order tensor with dimension $n_1 \times \cdots \times n_d$. Let $\mathbf Y=\mathbf \Theta + \mathcal E$, where $\mathbf \Theta=\mbox{E}(\mathbf Y)$ is our primary interest, and $\mathcal E$ is an $n_1 \times \cdots \times n_d$-dimensional random error. 
We assume that $\mathcal E$ has i.i.d elements with mean 0, variance $\sigma^2$ and a finite moment generating function at an open interval containing 0. For an arbitrary element $y_{i_1i_2\cdots i_d}$ of $\mathbf Y$, the $L_2$-loss function is 
\ba \label{l2loss}
l(\mathbf \Theta, y_{i_1i_2\cdots i_d})=(y_{i_1i_2\cdots i_d}- \theta_{i_1i_2\cdots i_d})^2,
\ea
where $\theta_{i_1i_2\cdots i_d}=\sum_{j=1}^r (p^1_{i_1j}+q^1_{i_1j}) (p^2_{i_2j}+q^2_{i_2j}) \cdots (p^d_{i_dj}+q^d_{i_dj})$ is the corresponding element in $\mathbf \Theta$ and is a function of $P$ and $Q$. Notice that each $l(\mathbf \Theta, y_{i_1i_2\cdots i_d})$ relies on $\mathbf \Theta$ only through $\theta_{i_1i_2\cdots i_d}$.
Since, in practice, the utilities are usually non-negative and finite, we assume that $\|(P,Q)\|_{\infty} \leq c_0$, where $c_0$ is a positive constant.
Let $J(\mathbf \Theta)$ be a non-negative penalty function. Then the overall criterion function is re-defined as
\be \label{criterion}
L(\mathbf \Theta|\mathbf Y)=\sum_{(i_1,\cdots,i_d) \in \Omega}l(\mathbf \Theta, y_{i_1i_2\cdots i_d})+\lambda_{|\Omega|}J(\mathbf \Theta), \mbox{ for }\mathbf \Theta \in \mathcal{S},
\ee
where $\lambda_{|\Omega|}$ is the penalization coefficient and $\mathcal{S} \subseteq \mathbb{R}^{n_1\times \cdots \times n_d}$ is the parameter space for $\mathbf \Theta$.

{\color{black}
Let $\mathbf \Theta_0$ be the unique true parameter.
We assume that $\hat{\bfTheta}_{|\Omega|}$ is a sample estimator of $\mathbf \Theta_0$ satisfying:
\be \label{SIEVE}
L(\hat{\bfTheta}_{|\Omega|}|\mathbf Y) \le \inf_{\Theta \in \mathcal{S}} L(\bfTheta|\mathbf Y) +\tau_{|\Omega|},
\ee
where $\displaystyle\lim_{|\Omega| \rightarrow \infty} \tau_{|\Omega|}=0$. {\color{black} Note that $\hat{\bfTheta}_{|\Omega|}$ is identifiable by 
Proposition \ref{identifiability}. Condition (\ref{SIEVE}) implies that $\hat{\bfTheta}_{|\Omega|}$ converges to a global minimizer of $L(\bfTheta|\mathbf Y)$ as $|\Omega| \rightarrow \infty$. However, the verification of (\ref{SIEVE}) 
could still be challenging in that $L$ is non-convex in $\bfTheta$ given missing values 
associated with $\mathbf{Y}$. Nevertheless, for a finite sample, condition (\ref{SIEVE}) does not impose any restrictions on the existence of a global minimum, and $\hat{\bfTheta}_{|\Omega|}$ is not required to be a global minimizer even if one exists. In practice, one could use a good initial point to sufficiently reduce the 
value of $L$, as discussed in Section 4.2.2, which would lead to more satisfactory 
numerical results.} 
}
}

Let $l_{\Delta}(\mathbf \Theta|\cdot)=l(\mathbf \Theta, \cdot) - l(\mathbf \Theta_0, \cdot)$ be the loss difference, and
\be \label{K}
\displaystyle K(\mathbf \Theta_0, \mathbf \Theta)=\frac{1}{n_1\cdots n_d}\sum_{i_1=1}^{n_1}\cdots\sum_{i_d=1}^{n_d}\mbox{E}\left\{l_{\Delta}(\mathbf \Theta, y_{i_1i_2\cdots i_d})\right\}
\ee
 be the expected loss difference. Since $\mathbf \Theta_0$ is the unique true parameter, we have $K(\mathbf \Theta_0, \mathbf \Theta) \ge 0$ for all $\mathbf \Theta \in \mathcal{S}$ and $K=0$ only if $\mathbf \Theta=\mathbf \Theta_0$. We define the distance between $\mathbf \Theta$ and $\mathbf \Theta_0$ as $\rho(\mathbf \Theta_0, \mathbf \Theta)=K^{1/2}(\mathbf \Theta_0, \mathbf \Theta)$, and let 
\be \label{V}
\displaystyle V(\mathbf \Theta_0, \mathbf \Theta)=\frac{1}{n_1\cdots n_d}\sum_{i_1=1}^{n_1}\cdots\sum_{i_d=1}^{n_d}\mbox{Var}\left\{l_{\Delta}(\mathbf \Theta, y_{i_1i_2\cdots i_d})\right\}.
\ee
Then, under the $L_2$-loss, we have $K(\mathbf \Theta_0, \mathbf \Theta)=\frac{1}{n_1\cdots n_d}\|\mathbf \Theta_0- \mathbf \Theta\|^2$ and $V(\mathbf \Theta_0, \mathbf \Theta)=\frac{4\sigma^2}{n_1\cdots n_d} \|\mathbf \Theta_0- \mathbf \Theta\|^2$, where $\|\cdot\|$ stands for the Euclidean norm of the vectorized tensor.


\begin{theorem} \label{l2convergence}
\textcolor{black}{Suppose $\hat{\mathbf \Theta}_{|\Omega|}$ is a sample estimator satisfying (\ref{SIEVE}).} Then we have:
\[ P(\rho(\hat{\mathbf \Theta}_{|\Omega|},\mathbf \Theta_0) \ge \eta_{|\Omega|}) \leq 7\exp(-c_1|\Omega| \eta_{|\Omega|}^2),\]
where $c_1 \ge 0$ is a constant, $\eta_{|\Omega|}=\max (\varepsilon_{|\Omega|},\lambda^{1/2}_{|\Omega|})$, and $\varepsilon_{|\Omega|} \sim \frac{1}{{|\Omega|}^{1/2}}$ is the best possible rate achieved when $\lambda_{|\Omega|} \sim \varepsilon^2_{|\Omega|}$.
\end{theorem}

Theorem 1 states that if the penalty term shrinks to zero at a rate no slower than 
the rate $\varepsilon^2_{|\Omega|}$ as the sample size tends to infinity, then the proposed method can achieve the convergence rate of $\frac{1}{{|\Omega|}^{1/2}}$, the same rate as the maximum likelihood estimator. In addition, the development of the convergence rate is under the $L_2$ distance, which, as a special case of the Kullback-Leiber information, is stronger than the commonly used Hellinger distance \citep{shen1998method}.

\subsubsection{\textcolor{black}{General Asymptotic Properties}}

 Next we develop the estimation consistency under more general settings. For each element of $\mathbf Y$, we assume $$\mbox{E}(y_{i_1 \cdots i_d})=\nu(\theta_{i_1 \cdots i_d}),$$ where $\nu(\cdot)$ is a mean function. For example, when $\mathbf Y$ is binary, we might adopt the logistic link $\nu(\theta)=\frac{\exp(\theta)}{1+\exp(\theta)}$, and when $\mathbf Y$ is ordinal, we have $\nu(\theta)=\exp(\theta)$. We also assume $\mbox{Var}(y_{i_1 \cdots i_d}) < \infty$.

In this general setting, $l(\cdot,\cdot)$ in (\ref{criterion}) is not necessarily an $L_2$-loss function. Let $K(\mathbf \Theta_0, \mathbf \Theta)$ and $V(\mathbf \Theta_0, \mathbf \Theta)$ be defined as in (\ref{K}) and (\ref{V}), respectively, and let $\rho(\mathbf \Theta_0, \mathbf \Theta)=K^{1/2}(\mathbf \Theta_0, \mathbf \Theta)$. Here $K(\cdot, \cdot)$ reduces to the Kullback-Leiber pseudo-distance if $l(\cdot, \cdot)$ corresponds to a log-likelihood.
Let $W_p^{\alpha}[a,b]^{n_1\times \cdots \times n_d}$ be a Sobolev space with finite $L_p$-norm, where $a$ and $b$ are some constants and $\alpha$ is the parameter associated with the degree of smoothness of functions \citep{devore1993constructive}.

\begin{assumption} \label{smoothness}
For each $y_{i_1 \cdots i_d}$, suppose
$$|l(\mathbf \Theta_0,y_{i_1 \cdots i_d})-l(\mathbf \Theta,y_{i_1 \cdots i_d})| \le g(y_{i_1 \cdots i_d}) \|\mathbf \Theta_0-\mathbf \Theta\|,$$
where $g(\cdot)$ satisfies $\emph{E}[\exp\{{t_0 g(y_{i_1 \cdots i_d})}\}] \le c_2 < \infty$, for a constant $c_2$ and some constants $t_0$ around 0. In particular, there exists a constant $c_2'>0$, such that $\mbox{E}\{g^2(y_{i_1,\ldots,i_d})\}\le c_2'$ for all $y_{i_1 \cdots i_d}$'s.
\end{assumption}

\begin{assumption} \label{ball}
Suppose there exist $\delta>0$ and $\beta \in [0,1)$, such that for a $\delta$-ball centered at $\mathbf \Theta_0$, we have $\rho(\mathbf \Theta_0, \mathbf \Theta) \ge c_3 \|\mathbf \Theta_0-\mathbf \Theta\|^{\frac{1}{1+\beta}}$, where $c_3 \ge 0$ is a constant.
\end{assumption}

Assumption \ref{ball} indicates that in a neighborhood of the true parameter $\mathbf \Theta_0$, the distance $\rho(\mathbf \Theta_0,\cdot)$ is no smaller than the Euclidean distance up to a certain order. On the contrary, if $\rho(\mathbf \Theta_0,\cdot)$ is dominated by the Euclidean distance for all neighborhoods of $\mathbf \Theta_0$, then the convergence result under $\rho(\cdot,\cdot)$ can be shown similar to the proof of Theorem \ref{l2convergence}. 


\begin{theorem} \label{generalconvergence}
Let $\hat{\mathbf \Theta}_{|\Omega|}$ be a sample estimator satisfying (\ref{SIEVE}). Assume that $l_{\Delta} \in W_p^{\alpha}[a,b]^{n_1\times \cdots \times n_d}$, where $p > 2$, and that Assumptions \ref{smoothness} and \ref{ball} hold. Then:
\[ P(\rho(\hat{\mathbf \Theta}_{|\Omega|},\mathbf \Theta_0) \ge \eta_{|\Omega|}) \leq 7\exp(-c_4|\Omega| \eta_{|\Omega|}^2),\]
where $c_4 \ge 0$ is a constant, and $\eta_{|\Omega|}=\max (\varepsilon_{|\Omega|},\lambda^{1/2}_{|\Omega|})$ with 
\[ \varepsilon_{|\Omega|} \sim
  \begin{cases}
\left(\frac{1}{{|\Omega|}^{1/2}}\right)^{\frac{2\omega}{2\omega+1}}       & \emph{if } \omega>\frac{1}{2}\\
\left(\frac{1}{{|\Omega|}^{1/2}}\right)^{\omega}  & \emph{if } \omega\leq\frac{1}{2}
  \end{cases}
\]
being the best possible rate, which can be achieved when $\lambda_{|\Omega|} \sim \varepsilon^2_{|\Omega|}$. Here $\omega=\alpha/\gamma$, and $\gamma=\sum_{k=1}^d (n_k+m_k)r$ is the total number of parameters.
\end{theorem}

The assumption of $p>2$ can be relaxed to $p \ge 2$ if $p\omega$ does not go to 0 as $|\Omega| \rightarrow \infty$. Notice that the convergence rate in Theorem \ref{generalconvergence} becomes $\varepsilon_{|\Omega|}\sim\frac{1}{{|\Omega|}^{1/2}}$ if $\omega=\infty$, which is the convergence rate of the maximum likelihood estimator achieved in Theorem \ref{l2convergence}.

\section{Simulation Studies} \label{Simulation Studies}

In this section, we perform simulations to compare the proposed method (REM) with five competing tensor factorization methods. Three methods are existing methods, namely, Bayesian probabilistic tensor factorization \citep[BPTF;][]{xiong2010temporal}, the factorization machine \citep[libFM;][]{rendle2012factorization}, and the Gaussian process factorization machine \citep[GPFM;][]{nguyen2014gaussian}.\footnote{The codes can be obtained from \url{https://www.cs.cmu.edu/~lxiong/bptf/bptf.html}, \url{http://www.libfm.org/}, and \url{http://trungngv.github.io/gpfm/}, respectively.}
{\color{black}
Since subgroup information is available, another naive but effective tensor factorization method is to conduct CPD for each subgroup separately, and combine the final result. This method is referred to as the groupwise Canonical Polyadic Decomposition (GCPD).
}
\textcolor{black}{In addition, we also investigate the performance of matrix factorization (MF) under the tensor framework, which is the misspecified proposed method with $d=2$ and ignoring contextual information.}

\subsection{The ``Cold-Start'' Problem} \label{cs problem}
The first simulation study is designed to compare the performance of each method under various severity levels of the ``cold-start'' problem. Specifically, we consider a third-order tensor with \textit{user}, \textit{item} and one \textit{contextual variable}. We set the number of users $n_1=400$, the number of items $n_2=1100$ and the number of contexts $n_3=9$. We assume that the users, items and contexts are from $m_1=10$, $m_2=11$ and $m_3=3$ subgroups, respectively, and assume the number of latent factors $r=3$.
We generate each latent factor $\mathbf p_{i_k}^k \stackrel{iid}{\sim} N(0,\mathbf I_r)$ for $i_k=1,\ldots,n_k$, and $k=1,2,3$. To distinguish different subgroups, we set the nested factors as a simple ordered sequence, where $\mathbf q^1_{(u_1)}=(-5.5+{u_1})\mathbf{1}_r$, $\mathbf q^2_{(u_2)}=(-3.6+0.6{u_2})\mathbf{1}_r$, and $\mathbf q^3_{(u_3)}=(-4+2{u_3})\mathbf{1}_r$ for $u_k=1,\ldots,m_k$. Users, items and contexts are evenly assigned to each subgroup.

For each simulation, we generate $N=n_1n_2n_3(1-\pi_0)$ entries out of the entire tensor, where 
$\pi_0=80\%$, 95\% or 99\% are the missing percentages. Furthermore, we use $\phi_{cs}$ to measure the severity of the ``cold-start'' problem, and $\phi_{cs}=30\%$, 60\% or 95\% represents the proportion of the testing data whose utilities are about new items and are not available from the training set. Each utility is generated by $y_{i_1i_2i_3}=\sum_{j=1}^r (p^1_{i_1j}+q^1_{i_1j}) (p^2_{i_2j}+q^2_{i_2j})(p^3_{i_3j}+q^3_{i_3j})/3+\varepsilon,$ where $\varepsilon \sim N(0,1)$ is the random error.

For all methods, we assume that $r=3$ is known. 
{\color{black}
For REM and GCPD, we also assume that the subgroup memberships are correctly specified, and the tuning parameter $\lambda$ is pre-selected from grid points ranging from 1 to 11. Since the subgroup structure is assumed to be known, these two methods have more advantage and are expected to have better performance. 
}
For MF, we assume the same setting as the proposed method. For BPTF, we choose the number of Gibbs samples to be 50 and keep the remaining parameters by their default choices. For libFM, we use their default setting, and for GPFM, we select the radial basis function kernel with noise being equal to 10 and the standard deviation of latent variables being equal to 1. All methods are replicated by 200 simulation runs.

Table \ref{sim1} provides the performance of each method based on the root mean square error (RMSE) and the mean absolute error (MAE), where the MAE is defined as $\frac{1}{|\Omega|}\sum_{\Omega}|y_{i_1\cdots i_d}-\hat{y}_{i_1\cdots i_d}|$. We observe that most methods perform worse when either the missing percentage or the severity of the ``cold-start'' problem increases. In contrast, the proposed method is relatively robust against these changes. Specifically, REM performs the best across all settings, especially in the worst setting when the missing percentage is 99\% and the proportion of new items reaches 95\%; that is, 95\% of the items are new and are not available from the training set. In this scenario, the proposed method is at least 100\% better than other methods in terms of both the RMSE and the MAE. 



\subsection{High-Order Tensors} \label{hotensor}

We design the second simulation study to evaluate the performance of the proposed method under fourth-order tensors. 
Specifically, we let the number of users and the number of items be the same, namely, 500 or 1000, and the number of contexts be 4 for the two contextual variables. \textcolor{black}{Furthermore, we allow 10 subgroups for users and for items, and 2 subgroups for each of the two contextual variables. We also assume that the number of members in each subgroup is the same.} The number of latent factors is $r=3$, and each latent factor is generated from an i.i.d standard normal distribution. The nested factors for users and items are $(-5.5+{u})\mathbf{1}_r$ for $u=1,\ldots,10$, and the nested factors for the two contextual variables are $-0.25\cdot\mathbf{1}_r$ and $0.25\cdot\mathbf{1}_r$ corresponding to the two subgroups. Each utility is generated as $y_{i_1i_2i_3i_4}=\sum_{j=1}^r (p^1_{i_1j}+q^1_{i_1j})(p^2_{i_2j}+q^2_{i_2j})(p^3_{i_3j}+q^3_{i_3j})(p^4_{i_4j}+q^4_{i_4j})/4+\varepsilon,$ where $\varepsilon \sim N(0,1)$ is the random error.

For each replication in the tensor, we assume that the missing percentage $\pi_0=95\%, 97\%$ or $99\%$, \textcolor{black}{corresponding to the high-missing situation for high-order tensors}. Furthermore, we assume that $30\%$ of the items are not available in the training set, that is, the ``cold-start'' severity level is fixed. The tuning parameter selection for each method is the same as in simulation study 1. All methods are replicated by 200 simulations. REM, GCPD, libFM and GPFM are able to utilize the fourth order of the tensor, while MF and BPTF use up to the second and the third order, respectively.

Table \ref{sim2} provides the comparisons of the proposed method and other methods under the fourth-order tensor setting. It is clear that REM has the overall best performance in terms of both RMSEs and MAEs. \textcolor{black}{For the two misspecified methods, namely the MF and BPTF, their performances are similar to each other and close to the unreported grand mean imputation, although BPTF has better performance when the missing percentage is low.} For the three correctly-specified competing methods, namely GCPD, libFM and GPFM, their performances are not significantly better than the misspecified methods, although they utilize all tensor information. In contrast, the proposed method provides much smaller RMSEs and MAEs in all settings. For example, when the missing percentage is 99\% and the number of users and items are both equal to 1000, the existing methods perform mostly similar to that of the grand mean imputation, producing RMSEs slightly above 4 and MAEs around 2.6. That is, these methods do not utilize subject-specific information effectively. In contrast, the proposed method is able to improve on both the RMSE and the MAE by more than 100\%.

\section{IRI Marketing Data} \label{realdata}
In this section, we apply the proposed method to IRI Marketing Data \citep{bronnenberg2008database}. \textcolor{black}{The data contain 116.3 million observations of average sales volumes collected from 2447 grocery stores on 161,114 products from 2001 to 2011.} There are 30 promotion strategies for various products to attract consumers. Each observation consists of a store ID, a product ID, a promotion strategy and the corresponding average sales volume. The 2447 stores are selected from 47 markets across the United States, where demographic information within two miles of each store is collected as well. The 161,114 products include all items sold from these stores during the 11-year period. These products can be classified into 31 categories, including beer, coffee, frozen pizza, paper towels, etc. \textcolor{black}{The 30 promotion strategies are combinations of 5 advertisement features, 3 types of merchandise display, and an indicator on whether the product has a price reduction of more than 5\%.} The data have a $99\%$ missing rate after being re-organized into a third-order tensor by store, product and promotion.

\textcolor{black}{The goal of our study is to predict the average sales volume of each product from each store, and the average sales volume when a particular promotion strategy is applied. Through this prediction procedure, we are able to potentially recommend the most profitable products for each store, and evaluate how each specific promotion strategy plays a role for each product sales.}

The data are randomly split into a $50\%$ training set, a $25\%$ validation set and a $25\%$ testing set. The random split is replicated 50 times. Sales volumes from each category of product are standardized before analysis to avoid large differences of sales volumes from different categories. We compare the proposed method with the existing methods listed in Section \ref{cs problem}. In addition, we also compare it to the grand mean imputation where all missing elements in the tensor are imputed by the mean of the observed values. Tuning parameters for each method are selected from a wide range of grid points to minimize the RMSE on the validation set. \textcolor{black}{Most methods require that the number of latent factors $r \ge 15$ in order to capture a majority of the data variation.} For the proposed method, we classify stores, products and promotion strategies into subgroups based on their geographical locations (the first digit of the zipcode), product categories, and whether a price reduction is applied, respectively. 

Since the data are standardized, the grand mean imputation returns an RMSE close to 1, which can be regarded as a benchmark basis for comparison. Table \ref{real data} indicates that the proposed REM has the best performance in terms of both RMSE and MAE. Specifically, REM improves the RMSE of the MF by the largest percentage, which demonstrates the great advantage of incorporating contextual information through tensor structure. In addition, the RMSE of the proposed method is less than that of BPTF and libFM, illustrating that REM has better prediction accuracy among the competing tensor factorization methods. GPFM does not converge due to the high missing rate and the large number of parameters involved. Meanwhile, BPTF has nearly the same MAE as the proposed method, but produces a larger RMSE, indicating that BPTF's performance is possibly better than the proposed method for certain subjects, but has inferior performance for the rest of the subjects. 

{\color{black}
Furthermore, the naive method GCPD has the second best performance. This might be explained by the following two reasons. One is that the GCPD utilizes additional subgroup information as does the REM, which is not applicable to the other competing methods. The other is that, unlike in the simulation studies,
the size of the IRI data is large enough so that the number of observations within each subgroup is sufficient to make good estimations.}

Most of the numerical studies are implemented on Dell C8220 computing sleds equipped with two 10-core Intel Xeon E5-2670V2 processors and 64GB RAM. The running time for each method is provided in Table \ref{real data}. Notice that BPTF's method also requires large memory storage due to the high demand of the Gibbs samples. Reducing the number of Gibbs samples may lead to more efficient computation but less accurate predictions.


\section{Discussion} \label{Discussion}

In this article, we propose a new tensor-based recommender system which makes recommendations through incorporating contextual information. A unique contribution of our method is that we achieve tensor completion through utilizing an additional layer of nested factors, in addition to applying the latent factors as in the commonly used Canonical Polyadic Decomposition. The nested factors incorporate subject-subject dependency, which are estimated based on subgrouping. This leads to a significant advantage in solving the ``cold-start'' problem effectively. That is, for a new subject, information from other subgroup members can be borrowed to make more accurate predictions even though the subject's own information is not collected sufficiently. In addition, the proposed method is able to address high-order tensors which are beyond the third-order. Existing methods are not effective in handling the high-order tensor problem due to the high computational cost and the high missing rate. We propose a new algorithm that integrates maximum block improvement into the blockwise coordinate descent algorithm, which avoids operating high-order tensors directly and therefore achieves scalable computation. Moreover, the proposed nested factors borrow information from all members in the same subgroup and hence accommodate a high missing rate.

The proposed method also shows excellent numerical performance and theoretical properties. In Section \ref{realdata}, the proposed method is applied to IRI marketing data which consists of 116 million observations. The proposed method improves prediction accuracy compared to existing methods with relatively small computational cost. 
{\color{black}
In theory, we demonstrate the convergence properties of the proposed algorithm, which {\color{black}converges to a stationary point from an arbitrary initial point}, and local convergence to a local minimum with linear convergence rate. The estimated parameter achieves asymptotic consistency under the $L_2$-loss function and other more general circumstances.
}

\clearpage
\begin{table}
\begin{footnotesize}
\caption{The proposed method (REM) is compared to matrix factorization (MF), groupwise Canonical Polyadic Decomposition (GCPD), Bayesian probabilistic tensor factorization \citep[BPTF;][]{xiong2010temporal}, factorization machine \citep[libFM;][]{rendle2012factorization}, and Gaussian process factorization machine \citep[GPFM;][]{nguyen2014gaussian} on simulated third-order tensors with different missing rates and different degrees of the ``cold-start'' problem; the RMSE and the MAE are provided with standard error in each parenthesis; $\pi_0$ and $\phi_{cs}$ represent the missing rate and the severity of the ``cold-start'' problem, respectively; and the simulation results are based on 200 replications.}
\begin{center}
\begin{tabular}{cccccc}
\hline
\hline

			&				&		&$\phi_{cs}=0.3$&$\phi_{cs}=0.6$&$\phi_{cs}=0.95$\\
 \hline
$\pi_0=80\%$	&REM 		&RMSE	&4.586 (0.562)	&6.467 (0.823)	&8.228 (1.052)\\
			&				&MAE	&2.210 (0.214)	&3.606 (0.432)	&5.292 (0.687)\\
			&MF			 	&RMSE	&10.307 (1.012)&10.279 (1.020)&10.373 (1.062)\\
			&				&MAE	&6.789 (0.658)	&6.783 (0.673)	&6.822 (0.700)\\	
			&GCPD		&RMSE	&5.892 (0.754)	&8.083 (0.851)	&9.949 (0.974)\\
			&				&MAE	&2.781 (0.226)	&4.467 (0.421)	&6.389 (0.628)\\
			&BPTF		&RMSE	&5.793 (0.590)	&8.142 (0.837)	&10.192 (1.050)\\
			&				&MAE	&2.645 (0.222)	&4.442 (0.438)	&6.518 (0.685)\\
			&libFM		&RMSE	&10.361 (1.070)&10.373 (1.070)&10.355 (1.073)\\
			&				&MAE	&6.778 (0.707)	&6.786 (0.713)	&6.768 (0.718)\\
			&GPFM		&RMSE	&9.017 (1.391)&9.852 (1.894)	&11.068 (1.406)\\
			&				&MAE	&5.215 (1.131)	&5.624 (0.761)	&6.992 (0.814)\\
 \hline
 $\pi_0=95\%$	&REM	&RMSE	&3.322 (0.510)	&4.658 (0.773)	&6.082 (1.055)\\
			&				&MAE	&1.760 (0.196)	&2.702 (0.401)	&3.943 (0.674)\\
			&MF			&RMSE	&10.768 (1.179)&10.728 (1.148)&10.656 (1.160)\\
			&				&MAE	&7.055 (0.761)	&7.045 (0.751)	&6.992 (0.761)\\
			&GCPD		&RMSE	&13.489 (2.059)	&12.247 (1.632)	&10.461 (1.153)\\
			&				&MAE	&6.213 (0.826)	&6.386 (0.767)	&6.638 (0.701)\\
			&BPTF	&RMSE	&5.847 (0.696)	&8.173 (0.881)	&10.197 (1.050)\\
			&				&MAE	&2.736 (0.411)	&4.498 (0.507)	&6.524 (0.683)\\
			&libFM	&RMSE	&10.384 (1.074)&10.395 (1.071)&10.359 (1.076)\\
			&				&MAE	&6.789 (0.717)	&6.799 (0.716)	&6.769 (0.719)\\
			&GPFM	&RMSE	&10.397 (1.072)&10.406 (1.070)	&10.369 (1.078)\\
			&				&MAE	&6.812 (0.718)	&6.822 (0.717)	&6.792 (0.720)\\
 \hline
$\pi_0=99\%$	&REM		&RMSE	&3.361 (1.032)	&4.329 (1.379)	&4.988 (1.582)\\
			&				&MAE	&1.865 (0.414)	&2.588 (0.722)	&3.289 (1.021)\\
			&MF			 	&RMSE	&12.647 (1.940)&12.612 (1.883)&12.379 (1.706)\\
			&				&MAE	&8.016 (1.071)	&8.016 (1.050)	&7.903 (0.981)\\
			&GCPD		&RMSE	&10.925 (1.191)	&10.808 (1.124)	&10.283 (1.058)\\
			&				&MAE	&6.748 (0.755)	&6.808 (0.721)	&6.685 (0.691)\\
			&BPTF	&RMSE	&8.246 (2.829) &9.243 (1.737)	&10.268 (1.093)\\
			&				&MAE	&4.696 (1.917)	&5.537 (1.256)	&6.616 (0.723)\\
			&libFM		&RMSE	&10.387 (1.058)	&10.389 (1.058)&10.342 (1.075)\\
			&				&MAE	&6.790 (0.707)	&6.792 (0.713)	&6.759 (0.717)\\
			&GPFM	&RMSE	&10.450 (1.068)	&10.456 (1.072)&10.410 (1.084)\\
			&				&MAE	&6.874 (0.712)	&6.878 (0.722)	&6.845 (0.723)\\
\hline
\hline
\end{tabular}

\end{center}

\label{sim1}
\end{footnotesize}

\end{table}

\clearpage
\begin{table}
\begin{footnotesize}
\caption{The proposed method (REM) is compared to matrix factorization (MF), groupwise Canonical Polyadic Decomposition (GCPD), Bayesian probabilistic tensor factorization \citep[BPTF;][]{xiong2010temporal}, factorization machine \citep[libFM;][]{rendle2012factorization}, and Gaussian process factorization machine \citep[GPFM;][]{nguyen2014gaussian} on simulated fourth-order tensors with different missing rates and different number of users and items; the RMSE and the MAE are provided with standard error in each parenthesis; $\hat{d}$ is the assumed tensor order for each method with the true order being 4; $\pi_0$, $n_1$ and $n_2$ represent the missing rate, the number of users and the number of items, respectively; and the simulation results are based on 200 replications.}
\begin{center}
\begin{tabular}{ccccccc}
\hline
\hline

			&		&$\hat{d}$		&		&$n_1=n_2=500$&$n_1=n_2=1000$\\
 \hline
$\pi_0=95\%$	&REM	&$\hat{d}=4$		&RMSE	&1.427 (0.351)	&1.799 (0.570)	\\
			&		&		&MAE	&1.015 (0.138)	&1.148 (0.202)	\\
			&MF		&$\hat{d}=2$		&RMSE	&4.040 (1.206)	&4.099 (1.373)\\
			&		&		&MAE	&2.582 (0.679)	&2.601 (0.759)	\\	
			&GCPD&$\hat{d}=4$&RMSE	&4.691 (1.906)	&4.565 (1.807)		\\
			&		&		&MAE	&2.599 (0.732)	&2.395 (0.638)		\\
			&BPTF	&$\hat{d}=3$&RMSE	&3.944 (1.212)	&3.896 (1.322)		\\
			&		&		&MAE	&2.472 (0.650)	&2.454 (0.722)	\\
			&libFM	&$\hat{d}=4$		&RMSE	&4.057 (1.172)	&4.136 (1.423)	\\
			&		&		&MAE	&2.578 (0.655)	&2.585 (0.773)		\\
			&GPFM&$\hat{d}=4$&RMSE	&3.958 (1.103)	&3.052 (1.194)		\\
			&		&		&MAE	&2.533 (0.640)	&1.916 (0.602)		\\
 \hline
$\pi_0=97\%$	&REM	&$\hat{d}=4$		&RMSE	&1.512 (0.670)	&1.689 (0.524)	\\
			&		&		&MAE	&1.050 (0.227)	&1.108 (0.174)	\\
			&MF	&$\hat{d}=2$			 	&RMSE	&4.048 (1.210)	&4.103 (1.375)\\
			&		&		&MAE	&2.588 (0.682)	&2.605 (0.761)	\\	
			&GCPD&$\hat{d}=4$&RMSE	&4.640 (1.792)	&5.384 (1.876)		\\
			&		&		&MAE	&2.665 (0.795)	&2.779 (0.699)		\\
			&BPTF	&$\hat{d}=3$&RMSE	&4.047 (1.264)	&3.952 (1.345)		\\
			&		&		&MAE	&2.503 (0.662)	&2.473 (0.726)	\\
			&libFM	&$\hat{d}=4$		&RMSE	&4.087 (1.214)	&4.198 (1.371)	\\
			&		&		&MAE	&2.593 (0.677)	&2.648 (0.768)		\\
			&GPFM	&$\hat{d}=4$&RMSE	&4.115 (1.215)	&3.604 (1.128)		\\
			&		&		&MAE	&2.616 (0.682)	&2.341 (0.634)		\\
 \hline
$\pi_0=99\%$	&REM	&$\hat{d}=4$		&RMSE	&2.780 (1.915)	&1.880 (1.540)\\
			&		&		&MAE	&1.579 (0.742)	&1.187 (0.433)	\\
			&MF	&$\hat{d}=2$			 	&RMSE	&4.108 (1.233)	&4.124 (1.390)\\
			&		&		&MAE	&2.625 (0.692)	&2.621 (0.769)	\\	
			&GCPD&$\hat{d}=4$&RMSE	&4.168 (1.364)	&4.453 (1.680)		\\
			&		&		&MAE	&2.567 (0.718)	&2.611 (0.753)		\\
			&BPTF&$\hat{d}=3$&RMSE	&4.399 (1.477)	&4.174 (1.497)		\\
			&		&		&MAE	&2.620 (0.706)	&2.540 (0.754)	\\
			&libFM&$\hat{d}=4$	&RMSE	&4.124 (1.226)	&4.164 (1.399)	\\
			&		&		&MAE	&2.613 (0.678)	&2.624 (0.761)		\\
			&GPFM&$\hat{d}=4$&RMSE	&4.121 (1.221)	&4.248 (1.391)		\\
			&		&		&MAE	&2.619 (0.680)	&2.678 (0.778)		\\
\hline
\hline
\end{tabular}

\end{center}
\label{sim2}

\end{footnotesize}

\end{table}

\clearpage
\begin{table}
\begin{footnotesize}
\caption{The proposed method (REM) is compared to matrix factorization (MF), groupwise Canonical Polyadic Decomposition (GCPD), Bayesian probabilistic tensor factorization \citep[BPTF;][]{xiong2010temporal}, factorization machine \citep[libFM;][]{rendle2012factorization}, Gaussian process factorization machine \citep[GPFM;][]{nguyen2014gaussian}, and the grand mean imputation (GMI; as a reference level) on 116 million IRI marketing data points; GPFM is not included because it fails to converge; and the other results are based on 50 replications of random testing sets. In the first tabular, the comparison is made based on root mean square error (RMSE), mean absolute error (MAE) and computational time in hours (Comp. Time); and in the second tabular, the relative improvement of the proposed method over existing methods is provided.}
\begin{center}
\begin{tabular}{cccc}
\hline
\hline

					&RMSE		&MAE		&Comp. Time (hrs)\\
 \hline
			REM		&0.637 (0.009)&0.209 (0.001)	&3.9\\
			MF		&0.969 (0.010)&0.371 (0.007)	&0.7\\
			GCPD		&0.640 (0.010)&0.229 (0.001)	&5.4\\
			BPTF	&0.782 (0.155)&0.209 (0.001)	&8.4\\
			libFM	&0.705 (0.010)&0.236 (0.001)	&0.5\\
			GMI	&1.000 (0.006)&0.392 (4.483$\times 10^{-5}$)	&N/A\\
\hline
\hline
\end{tabular}

\vspace*{5mm}

\begin{tabular}{ccc}

\hline
\hline

					&RMSE		&MAE		\\
 \hline
			MF		&34.2\%		&43.7\%	\\
			GCPD		&0.5\%		&8.7\%	\\
			BPTF	&18.5\%		&0\%	\\
			libFM	&9.6\%		&11.4\%	\\
			GMI		&36.3\%		&46.7\%	\\
\hline
\hline
\end{tabular}

\end{center}
\label{real data}

\end{footnotesize}

\end{table}

\clearpage

\setcounter{equation}{0}
\def\theequation{S\arabic{equation}}

\section*{Appendix}

\section*{\textcolor{black}{Proof of Proposition \ref{identifiability}}}

Recall that $$L(P,Q|\mathbf{Y})=\sum_{(i_1,\cdots,i_d) \in \Omega} \left(y_{i_1\cdots i_d} - \hat{y}_{i_1\cdots i_d}\right)^2 +\lambda \sum_{k=1}^d \sum_{j=1}^r(\|\mathbf{p}_{\cdot j}^k\|_2^2+\|\mathbf{q}_{\cdot j}^k\|_2^2),$$
where $\hat{y}_{i_1\cdots i_d}=\sum_{j=1}^r (p^1_{i_1j}+q^1_{i_1j})(p^2_{i_2j}+q^2_{i_2j}) \cdots (p^d_{i_dj}+q^d_{i_dj})$.

Suppose there exist two minimizers of $L(\cdot,\cdot|\mathbf{Y})$, namely $(P,Q)$ and $(\tilde{P},\tilde{Q})$. Under the assumption that $\sum_{k=1}^d \mathcal{K}_{B^k} \ge 2r+(d-1)$,  it follows from
Theorem 3 of \cite{sidiropoulos2000uniqueness} that $(P,Q)$ and $(\tilde{P},\tilde{Q})$ are identical with the exception of scaling, permutation and addition.
Then based on Lemma \ref{invariant}, the $\hat{y}_{i_1\cdots i_d}$'s provided by $(P,Q)$ and $(\tilde{P},\tilde{Q})$ are identical.

Therefore, $L(P,Q|\mathbf{Y})=L(\tilde{P},\tilde{Q}|\mathbf{Y})$ implies that 
\be \label{identi}
\sum_{k=1}^d \sum_{j=1}^r(\|\mathbf{p}_{\cdot j}^k\|_2^2+\|\mathbf{q}_{\cdot j}^k\|_2^2)=\sum_{k=1}^d \sum_{j=1}^r(\|\tilde{\mathbf{p}}_{\cdot j}^k\|_2^2+\|\tilde{\mathbf{q}}_{\cdot j}^k\|_2^2).
\ee

For scaling, suppose there exist some $k_1, k_2=1,\ldots,d$, $k_1 \neq k_2$, such that $\tilde{\mathbf{p}}_{\cdot j}^{k_1}=\alpha_j \mathbf{p}_{\cdot j}^{k_1}$ and $\tilde{\mathbf{q}}_{\cdot j}^{k_1}=\alpha_j \mathbf{q}_{\cdot j}^{k_1}$, whereas $\tilde{\mathbf{p}}_{\cdot j}^{k_2}=\frac{1}{\alpha_j} \mathbf{p}_{\cdot j}^{k_2}$ and $\tilde{\mathbf{q}}_{\cdot j}^{k_2}=\frac{1}{\alpha_j} \mathbf{q}_{\cdot j}^{k_2}$ for a constant $\alpha_j >0$, $j=1,\ldots,r$. We have
\ba
&&\sum_{j=1}^r(\|\tilde{\mathbf{p}}_{\cdot j}^{k_1}\|_2^2+\|\tilde{\mathbf{q}}_{\cdot j}^{k_1}\|_2^2)+\sum_{j=1}^r(\|\tilde{\mathbf{p}}_{\cdot j}^{k_2}\|_2^2+\|\tilde{\mathbf{q}}_{\cdot j}^{k_2}\|_2^2) \\
&=& \sum_{j=1}^r\alpha_j^2(\|\mathbf{p}_{\cdot j}^{k_1}\|_2^2+\|\mathbf{q}_{\cdot j}^{k_1}\|_2^2)+ \sum_{j=1}^r\frac{1}{\alpha_j^2}(\|\mathbf{p}_{\cdot j}^{k_2}\|_2^2+\|\mathbf{q}_{\cdot j}^{k_2}\|_2^2).
\ea
Then (\ref{identi}) implies that $\alpha_j=1$ almost surely, $j=1,\ldots,r$.

For addition, similarly, suppose $\tilde{\mathbf{p}}_{\cdot j}^k=\mathbf{p}_{\cdot j}^k+\bfdelta_j^k$ and $\tilde{\mathbf{q}}_{\cdot j}^k=\mathbf{q}_{\cdot j}^k-\bfdelta_j^k$ for a constant vector $\bfdelta_j^k$, $k=1,\ldots,d$ and $j=1,\ldots,r$. Then (\ref{identi}) implies that
\ba
&&\sum_{k=1}^d \sum_{j=1}^r(\|\tilde{\mathbf{p}}_{\cdot j}^k\|_2^2+\|\tilde{\mathbf{q}}_{\cdot j}^k\|_2^2)\\
&=& \sum_{k=1}^d \sum_{j=1}^r\big(\|\mathbf{p}_{\cdot j}^k\|_2^2+\|\mathbf{q}_{\cdot j}^k\|_2^2+2\|\bfdelta_j^k\|_2^2+2(\mathbf{p}_{\cdot j}^k)'\bfdelta_j^k-2(\mathbf{q}_{\cdot j}^k)'\bfdelta_j^k\big)\\
&=& \sum_{k=1}^d \sum_{j=1}^r(\|\mathbf{p}_{\cdot j}^k\|_2^2+\|\mathbf{q}_{\cdot j}^k\|_2^2),
\ea
which indicates that $\bfdelta_j^k=\mathbf{0}$ almost surely, $k=1,\ldots,d$ and $j=1,\ldots,r$.

This concludes that $(P,Q)=(\tilde{P},\tilde{Q})$ almost surely, except for permutation.
\hfill\qedsymbol

\section*{\textcolor{black}{Proof of Lemma \ref{stationary}}}

Recall that 
$X^{(-k)}=(X^1,\ldots,X^{k-1},$ $X^{k+1},\ldots,X^d)$ for $X=P$ or $Q$, $k=1,\ldots,d$. 
For each $P^k$, let 
$$U^k(P^{(-k)},Q)= \mathop{\argmin}_{P^k}L(P^k|\mathbf Y,P^{(-k)},Q).$$
Here $U^k(P^{(-k)},Q)$ is the unique minimizer of $L(\cdot|\mathbf{Y})$ along the direction of $P^k$, because each of its rows is a solution of the ridge regression (\ref{estp}).
Similarly, we define $V^k(P,Q^{(-k)})$ as the unique minimizer of $L(\cdot|\mathbf{Y})$ along the direction of $Q^k$.

{\color{black} In this proof, let $P^k_t$ and $Q^k_t$ denote the estimated $P^k$ and $Q^k$ at the $t$-th attempted update instead of the $t$-th iteration, respectively. At each attempted update, only one block of $(P^1_{t},\ldots,P^d_{t},Q^1_{t},\ldots,Q^d_{t})$ is updated. For example, if $P^k_t$ is updated, then $P^{k_1}_{t+1}=P^{k_1}_t$ and $Q^{k_2}_{t+1}=Q^{k_2}_t$ remain unchanged, for $k_1 \neq k$ and $k_2=1,\ldots,d$. 
Recall that, at each iteration, the MBI algorithm retains only one update
with the largest improvement in terms of the objective function value.
In other words, only a subsequence of $\{(P^1_{t},\ldots,P^d_{t},Q^1_{t},\ldots,Q^d_{t})\}_{t \ge 1}$ is adopted by the algorithm and actually converges to the limit. That is,
\ba
\lim_{s \rightarrow \infty} (U^1_{t_s},\ldots,U^d_{t_s},V^1_{t_s},\ldots,V^d_{t_s}) = (\tilde{P}^1,\ldots,\tilde{P}^d,\tilde{Q}^1,\ldots,\tilde{Q}^d).
\ea
Therefore, we shall focus our attention on the convergence property of this 
subsequence.}

We assume that ($P^k$)'s and ($Q^k$)'s are updated iteratively, as described in Algorithm 1.
 Subsequently, similar to the proof of Theorem 3.1 of \cite{chen2012maximum}, we show that $(\tilde{P},\tilde{Q})$ is a blockwise local minimizer. On the one hand, suppose that $P^k$ is updated at the $(t_s+1)$-th iteration and $Q^{k'}$ is updated at the $(t_s+2)$-th iteration, where $k$ is not necessarily equal to $k'$. Then,
$$P^k_{t_s+1}=U^k(P^{(-k)}_{t_s},Q_{t_s}) \mbox{ and }Q^{k'}_{t_s+2}=V^{k'}(P_{t_s+1},Q^{(-k')}_{t_s+1}).$$
Hence,
\ba
&&L(P^1_{t_s},\ldots,P^{k-1}_{t_s},U^k(\tilde{P}^{(-k)},\tilde{Q}),P^{k+1}_{t_s},\ldots,P^d_{t_s},Q^1_{t_s},\ldots,Q^d_{t_s}|\mathbf Y)\\
&\ge& L(P^1_{t_s},\ldots,P^{k-1}_{t_s},U^k(P^{(-k)}_{t_s},Q_{t_s}),P^{k+1}_{t_s},\ldots,P^d_{t_s},Q^1_{t_s},\ldots,Q^d_{t_s}|\mathbf Y)\\
&=& L(P^1_{t_s+1},\ldots,P^d_{t_s+1},Q^1_{t_s+1},\ldots,Q^d_{t_s+1}|\mathbf Y)\\
&\ge& L(P^1_{t_s+1},\ldots,P^d_{t_s+1},Q^1_{t_s+1},\ldots,Q^{k'-1}_{t_s+1},Q^{k'}_{t_s+2},Q^{k'+1}_{t_s+1},\ldots,Q^d_{t_s+1}|\mathbf Y)\\
&=& L(P^1_{t_s+2},\ldots,P^d_{t_s+2},Q^1_{t_s+2},\ldots,Q^d_{t_s+2}|\mathbf Y)\\
&\ge& L(P^1_{t_{s+1}},\ldots,P^d_{t_{s+1}},Q^1_{t_{s+1}},\ldots,Q^d_{t_{s+1}}|\mathbf Y).
\ea
As $s \rightarrow \infty$, 
\ba
&& L(\tilde{P}^1,\ldots,\tilde{P}^{k-1},U^k(\tilde{P}^{(-k)},\tilde{Q}),\tilde{P}^{k+1},\ldots,\tilde{P}^d,\tilde{Q}^1,\ldots,\tilde{Q}^d|\mathbf Y)\\
&\ge&L(\tilde{P}^1,\ldots,\tilde{P}^d,\tilde{Q}^1,\ldots,\tilde{Q}^d|\mathbf Y).
\ea
On the other hand, by the definition of $U^k(\tilde{P}^{(-k)},\tilde{Q})$, we have 
\ba
&& L(\tilde{P}^1,\ldots,\tilde{P}^{k-1},U^k(\tilde{P}^{(-k)},\tilde{Q}),\tilde{P}^{k+1},\ldots,\tilde{P}^d,\tilde{Q}^1,\ldots,\tilde{Q}^d|\mathbf Y)\\
&\le&L(\tilde{P}^1,\ldots,\tilde{P}^d,\tilde{Q}^1,\ldots,\tilde{Q}^d|\mathbf Y).
\ea
Thus, the above inequality holds as an equality. In a similar way, we can show that the same equality holds for $V^k(\tilde{P},\tilde{Q}^{(-k)})$, $k=1,\ldots,d$.
Therefore, by definition, $(\tilde{P}^1,\ldots,\tilde{P}^d,\tilde{Q}^1,\ldots,\tilde{Q}^d)$ is a blockwise local minimizer.  
\hfill\qedsymbol


\section*{\textcolor{black}{Proof of Proposition \ref{gc}}}

It can be seen that the estimation of latent factors and nested factors, as in (\ref{estp}) and (\ref{estq}), is similar to that of ridge regression. For the $i_k$-th subject in the $k$-th mode, we define a hypothetical design matrix $Z_{i_k}^k=(z_{i_k,ij}^k)_{|\Omega_{i_k}^k| \times r}$, where $i$ represents the $i$-th element of the sorted set $\{(i_1,\ldots,i_{k-1},i_{k+1},\ldots,i_d): (i_1,\ldots,i_d) \in \Omega\}$ and $z_{i_k,ij}^k=\prod_{l \ne k}(p_{i_lj}^l+q_{i_lj}^l)$.
Thus, each $L^k(P^k)$ can be represented by
$$L^k(P^k)=\sum_{i_k=1}^{n_k}\left\{ \|\mathbf{y}_{i_k} - Z_{i_k}^k(\mathbf{p}_{i_k}^k+\mathbf{q}_{i_k}^k)\|^2+\lambda \|\mathbf{p}_{i_k}^k\|_2^2\right\}, \mbox{ }k=1,\ldots,d,$$
where $\mathbf{y}_{i_k}$ is a vector of all $y_{i_1 \cdots i_d}$'s whose $k$-th index equals $i_k$.
Furthermore, we have the $n_kr$-dimensional gradient
$\nabla L^k= (\frac{\partial L^k}{\partial \mathbf{p}_1^k},\ldots,\frac{\partial L^k}{\partial \mathbf{p}_{n_k}^k})$, where 
$$\frac{\partial L^k}{\partial \mathbf{p}_{i_k}^k}=-2\left\{\mathbf{y}_{i_k} - Z_{i_k}^k(\mathbf{p}_{i_k}^k+\mathbf{q}_{i_k}^k)\right\}' Z_{i_k}^k+2\lambda (\mathbf{p}_{i_k}^k)', \mbox{ }k=1,\ldots,d.$$
The representation of $L^{k+d}(Q^k)$, $k=1,\ldots,d$, can be deducted similarly.

Next, we calculate the Hessian matrix of $L^k(\cdot)$. The second partial derivatives are
\[   
\frac{\partial^2 L^k}{\partial \mathbf{p}_{i_k}^k \partial \mathbf{p}_{i_l}^k} = 
     \begin{cases}
       2\left\{(Z_{i_k}^k)' Z_{i_k}^k+\lambda I_r\right\} &\quad\mbox{if }k=l,\\
       0 &\quad\mbox{if } k \ne l, \\
     \end{cases}
\]
where $I_r$ is an $r$-dimensional identity matrix.
Therefore, the Hessian matrix of $L^k(\cdot)$ is 
$$H(L^k)=2\left\{\mbox{diag}\left((Z_1^k)' Z_1^k,\ldots,(Z_{n_k}^k)' Z_{n_k}^k\right)+\lambda I_{n_kr}\right\}.$$
It can be seen that $H(L^k)$ is positive definite for $\lambda>0$. 
Hence, each blockwise function $L^k(\cdot)$ is strongly convex. For fixed $(P^1,\ldots,P^d,Q^1,\ldots,Q^d)$ and all $X_1, X_2 \in \mathbb{R}^{n_k \times r}$ satisfying $L^k(X_1), L^k(X_2) \le L^k(P^k)$, this implies that 
\be \label{strongly convex}
L^k(X_1) \ge L^k(X_2) + \langle \nabla L^k(X_2), X_1-X_2 \rangle_F +\frac{\xi^k}{2} \|X_1-X_2\|_F^2, 
\ee
where $k=1,\ldots,2d$, $\xi^k$ is a constant, and $\xi^k \ge \varepsilon > 0$ for a small constant $\varepsilon$. Notice that both $L^k(\cdot)$ and $\xi^k$ may depend on all block coordinates of $L(\cdot|\mathbf Y)$ except for the $k$-th block.

Meanwhile, since the Hessian matrix $H(L^k)$ of $L^k(X)$ is no longer a function of $X$, we have
\begin{flalign} \label{Lip}
\begin{aligned}
\|\nabla L^k(X_1) - \nabla L^k(X_2)\|_F &= \|H\left(L^k(X_2)\right) \cdot \mbox{vec}(X_1-X_2)\|_2\\
& \le \zeta^k \|X_1-X_2\|_F
\end{aligned}
\end{flalign}
by Taylor expansion, where $\mbox{vec}(X)$ represents the vectorized $X$. This holds true for fixed $(P^1,\ldots,P^d,Q^1,\ldots,Q^d)$ and all $X_1, X_2 \in \mathbb{R}^{n_k \times r}$ satisfying $L^k(X_1), L^k(X_2) \le L^k(P^k)$.


Now we verify the primary descent condition \citep{absil2005convergence}. Recall that $\{(P_s,Q_s)\}_{s \ge 1}$ is a sequence of estimated parameters generated by Algorithm 1. Then similar to the proof of Theorem 2.3 in \cite{li2015convergence}, we have:
\ba
&&L(P_s,Q_s|\mathbf{Y})-L(P_{s+1},Q_{s+1}|\mathbf{Y}) \\
&=& \{L(P_s,Q_s|\mathbf{Y})-L(P_{s+1},Q_{s}|\mathbf{Y})\} + \{L(P_{s+1},Q_s|\mathbf{Y})-L(P_{s+1},Q_{s+1}|\mathbf{Y})\}\\
&=& \max_{k=1,\ldots,d} \{ L^k(P^k_{s}|Q^k_s)-L^k(P^k_{s+1}|Q^k_s) \} \\
&&+ \max_{k=1,\ldots,d} \{ L^k(Q^k_{s}|P^k_{s+1})-L^k(Q^k_{s+1}|P^k_{s+1}) \} \\
&\ge& \max_{k=1,\ldots,d} \left( \frac{\xi^k_s}{2} \|P^k_s-P^k_{s+1}\|^2_F \right) + \max_{k=1,\ldots,d} \left( \frac{\xi^{k+d}_s}{2} \|Q^k_s-Q^k_{s+1}\|^2_F \right),
\ea
where $L^k(P^k_{s+1}|Q^k_s)$ is the same as $L^k(P^k_{s+1})$, but to emphasize that $Q_s^k$ has not been updated. The inequality holds because of (\ref{strongly convex}) with $\nabla L^k(P^k_{s+1}|Q^k_s)=\mathbf{0}$ and $\nabla L^k(Q^k_{s+1}|P^k_{s+1}) =\mathbf{0}$.

Furthermore, let $\displaystyle \xi_s=\min_{k=1,\ldots,2d} \xi^k_s$. Then for any $l_1, l_2 =1,\ldots,d$,
\ba
&&L(P_s,Q_s|\mathbf{Y})-L(P_{s+1},Q_{s+1}|\mathbf{Y}) \\
&\ge& \frac{\xi_s}{2} \left( \|P^{l_1}_s-P^{l_1}_{s+1}\|_F \cdot \|P^{k_1}_s-P^{k_1}_{s+1}\|_F + \|Q^{l_2}_s-Q^{l_2}_{s+1}\|_F \cdot \|Q^{k_2}_s-Q^{k_2}_{s+1}\|_F \right)\\
&\ge& \frac{\xi_s}{2 \zeta_s} \Big( \max_{k=1,\ldots,d} \| \nabla L^{k}(P^{k}_s) \|_F \cdot \|P^{k_1}_s-P^{k_1}_{s+1}\|_F \\
&&+ \max_{k=1,\ldots,d} \| \nabla L^{k+d}(Q^{k}_s) \|_F \cdot \|Q^{k_2}_s-Q^{k_2}_{s+1}\|_F \Big),
\ea
where $k_1, k_2 \in \{1,2,\ldots,d\}$ are the two blocks that are actually updated in the $(s+1)$-th iteration corresponding to $P^{k_1}_{s+1}$ and $Q^{k_2}_{s+1}$, respectively, and the second inequality is resulted from (\ref{Lip}). 
Then by the triangle inequality,
\ba
 \|(P_s,Q_s) -(P_{s+1},Q_{s+1}) \|_F &=& \|(P^{k_1}_s-P^{k_1}_{s+1})+(Q^{k_2}_s-Q^{k_2}_{s+1}) \|_F\\
& \le&  \|(P^{k_1}_s-P^{k_1}_{s+1})\|_F + \|(Q^{k_2}_s-Q^{k_2}_{s+1}) \|_F.
 \ea
Assumption \ref{homo} states that the improvement over $P^k_s$ and $Q^k_s$ is of the same order. This implies that there exists a sequence of constants $\{c_s\}_{s \ge 1}$, such that $\displaystyle\min_{s \ge 1}c_s \ge \varepsilon >0$, and 
$$\min \left\{ \max_{k=1,\ldots,d} \| \nabla L^{k}(P^{k}_s) \|_F, \max_{k=1,\ldots,d} \| \nabla L^{k+d}(Q^{k}_s) \|_F \right\} \ge c_s \max_{k=1,\ldots,2d} \| \nabla L^k\|_F.$$
Therefore, by combining the two inequalities above, we have:
\ba
&&L(P_s,Q_s|\mathbf{Y})-L(P_{s+1},Q_{s+1}|\mathbf{Y}) \\
&\ge& \frac{c_s\xi_s}{2 \zeta_s} \max_{k=1,\ldots,2d} \| \nabla L^k\|_F \left( \|P^{k_1}_s-P^{k_1}_{s+1}\|_F + \|Q^{k_2}_s-Q^{k_2}_{s+1}\|_F \right) \\
&\ge& \frac{c_s\xi_s}{2 \zeta_s} \max_{k=1,\ldots,2d} \| \nabla L^k\|_F \cdot \|(P_s,Q_s) -(P_{s+1},Q_{s+1}) \|_F.
\ea
Finally, since 
$$\| \nabla L\|_F^2 =\sum_{k=1}^{2d} \| \nabla L^k\|_F^2 \le 2d \max_{k=1,\ldots,2d} \| \nabla L^k\|_F^2,$$
we have the primary descent condition:
\ba
L(P_s,Q_s|\mathbf{Y})-L(P_{s+1},Q_{s+1}|\mathbf{Y}) \ge \frac{c_s\xi_s}{2 \sqrt{2d} \zeta_s} \| \nabla L\|_F \cdot \|(P_s,Q_s) -(P_{s+1},Q_{s+1}) \|_F.
\ea
Since $c_s$ and $\xi_s$ are bounded below and $ \zeta_s$ is bounded above, there exists a $\bar{\sigma} >0$, such that $\frac{c_s\xi_s}{2 \sqrt{2d} \zeta_s} \ge \bar{\sigma} >0$ for all $s \ge 1$.

By applying Proposition \ref{identifiability} and the rearrangement method beneath Proposition \ref{identifiability}, the complementary descent condition \citep{absil2005convergence} is satisfied almost surely.

Furthermore, since $L(P,Q|\mathbf{Y})$ is analytic in the neighborhood of $(\tilde{P},\tilde{Q})$ by definition (\ref{pracloss}), the cluster point $(\tilde{P},\tilde{Q})$ satisfies the Lojasiewicz gradient inequality \citep{lojasiewicz1965ensembles}. That is,
for all $(P,Q)$'s in some neighborhood of $(\tilde{P},\tilde{Q})$, there exist constants $c>0$ and $\kappa \in (0,1)$, such that
$$\| \nabla L(P,Q|\mathbf{Y}) \|_F \ge c|L(P,Q|\mathbf{Y})-L(\tilde{P},\tilde{Q}|\mathbf{Y})|^{1-\kappa}.$$
Then Theorem 3.2 of \cite{absil2005convergence} and Theorem 2.2 of \cite{li2015convergence} hold. Therefore, the cluster point 
estimated from Algorithm 1 is a limit point.

Then, following Lemma \ref{stationary} and Section 2.3 of \cite{li2015convergence}, since $L(\cdot|\mathbf{Y})$ is differentiable and the parameter space of $L(\cdot|\mathbf{Y})$ is open, the limit point of Algorithm 1 is a stationary point and a blockwise local minimizer.
\hfill\qedsymbol

\section*{\textcolor{black}{Proof of Proposition \ref{lc}}}

Without loss of generality, we assume that $(\tilde{P},\tilde{Q})=\mathbf{0}$ and $L(\tilde{P},\tilde{Q}|\mathbf{Y})=0$. 
For a given $(P,Q)$, we define the domain for the blockwise function $L^k(\cdot)$ as
$$\mathcal{D}^k(P,Q)=\left\{(P^1,\ldots,P^{k-1})\right\} \times \mathbb{R}^{n_k \times r} \times \left\{(P^{k+1},\ldots,P^{d},Q^{1},\ldots,Q^{d})\right\},$$
and
$$\mathcal{D}^{k+d}(P,Q)=\left\{(P^1,\ldots,P^{d},Q^{1},\ldots,Q^{k-1})\right\} \times \mathbb{R}^{n_k \times r} \times \left\{(Q^{k+1},\ldots,Q^{d})\right\},$$
for $k=1,\ldots,d$. Then $\cup_{k=1}^d \mathcal{D}^k(P,Q)$ and $\cup_{k=d+1}^{2d} \mathcal{D}^k(P,Q)$ represent the two search crosses for updating $P$ and $Q$, respectively. 

Without loss of generality, we let the neighborhood $\mathcal{V}$ be a ball based on the energy norm and centered at $(\tilde{P},\tilde{Q})=\mathbf{0}$. Since $(\tilde{P},\tilde{Q})$ is a strict local minimizer, we can set $\mathcal{V}$ small enough such that it is convex and $\{(P_s,Q_s)\}_{s \ge s_0} \subset \mathcal{V}$ exists for an $s_0 \ge 0$. For the $(s+1)$-th iteration of Algorithm 1, the search domain is defined as 
$$\mathcal{D}_{s+1}=\mathcal{V} \cap \left(\cup_{k=1}^d \mathcal{D}^k(P_s,Q_s)\cup_{k=d+1}^{2d} \mathcal{D}^k(P_{s+1},Q_s)\right).$$
Then by definition, $\displaystyle(P_{s+1},Q_{s+1}) =\mathop{\arg\min}_{\mathcal{D}_{s+1}} L(P,Q|\mathbf{Y})$ exists and is in $\mathcal{V}$.

Let $\displaystyle\bx_{s+1} = \mathop{\arg\min}_{\mathcal{D}_{s+1}} \|(P,Q)\|_E$ be a vector estimated from a block coordinate descent method that has the same block directions as Algorithm 1. It is known that $\{\bx_s\}_{s \ge 1}$ converges linearly in the energy norm, e.g., under the almost-cyclic rule or the Gauss-Southwell rule \citep{luo1992convergence}. Then there exists $\mu_1 \in [0,1)$, such that for a large $s$, we have
\be \label{rhs}
\|\bx_{s+1}\|_E \le \mu_1  \|(P_s,Q_s)\|_E.
\ee
Meanwhile, if we re-arrange $\bx_{s+1}$ as in the format of $(P,Q)$, then by the definition of $(P_{s+1},Q_{s+1})$, 
$$L(P_{s+1},Q_{s+1}|\mathbf Y) \le L(\bx_{s+1}|\mathbf Y).$$
Therefore, by Lemma 3.2 of \cite{li2015convergence}, for a small $\varepsilon >0$, there exists $\delta>0$, such that $\mathcal{V} \subset \mathcal{B}(\delta)$, where $ \mathcal{B}(\delta)$ is a ball centered at $\mathbf{0}$ with energy-norm-based radius equal to $\delta$. Then 
\be \label{lhs}
\|(P_{s+1},Q_{s+1})\|_E \le (1+\varepsilon)\|\bx_{s+1}\|_E.
\ee
Define $\mu = \mu_1(1+\varepsilon)$. Let $\varepsilon$ be small enough such that $\mu<1$. Then by (\ref{rhs}) and (\ref{lhs}):
$$\|(P_{s+1},Q_{s+1})\|_E \le \mu \|(P_s,Q_s)\|_E.$$
\hfill\qedsymbol

\section*{Proof of Theorem \ref{l2convergence}}

Let $A(k_1,k_2)=\{\mathbf \Theta \in \mathcal{S}: k_1 \leq \rho(\mathbf \Theta_0, \mathbf \Theta) \leq 2k_1, J(\mathbf \Theta) \leq k_2\}$, and $\mathcal{F}(k_1,k_2)=\{l_{\Delta}(\mathbf \Theta|\cdot): \mathbf \Theta \in A(k_1,k_2)\}$.

We first verify several conditions of Corollary 2 in \cite{shen1998method}. By definition, we have $\sup_{A(k_1,k_2)} V(\mathbf\Theta_0,\mathbf\Theta) \le c_8 k_1^2 =c_8 k_1^2\{1+(k_1^2+k_2)^{\beta_1}\}$, and hence $\beta_1=0$. In the rest of this section, all $c_i$'s with $i \in \mathbb{N}$ are assumed to be non-negative constants.

For a given $y_{i_1 \cdots i_d}$, we have $|\theta_{0,i_1 \cdots i_d}-\theta_{i_1 \cdots i_d}| \le \frac{1}{4\sigma^2} \mbox{Var}\{l(\mathbf\Theta,y_{i_1 \cdots i_d})-l(\mathbf\Theta_0,y_{i_1 \cdots i_d})\}$ for $i_k=1,\ldots,n_k$, $k=1,\ldots,d$.
Furthermore, $$|l(\mathbf\Theta,y_{i_1 \cdots i_d})-l(\mathbf\Theta_0,y_{i_1 \cdots i_d})|=|\theta_{0,i_1 \cdots i_d}-\theta_{i_1 \cdots i_d}| \cdot |2y_{i_1 \cdots i_d}-\theta_{0,i_1 \cdots i_d}-\theta_{i_1 \cdots i_d}|.$$ Define a new random variable $w=|2y_{i_1 \cdots i_d}-\theta_{0,i_1 \cdots i_d}-\theta_{i_1 \cdots i_d}|$, then we have $\mbox{E}\{\exp(t_0w)\} < \infty$ for $t_0$ at an open interval containing 0.

Now we verify that for a constant $c_9>0$, we have $\sup_{A(k_1,k_2)} \|\mathbf\Theta_0-\mathbf\Theta\|_{\sup} \le c_9 (k_1^2+k_2)^{\beta_2}$ for $\beta_2 \in [0,1)$. Define $f_0=f_0(P,Q)=\mathbf\Theta-\mathbf\Theta_0$. Recall that $\|(P,Q)\|_{\infty} \le c_0$ and $\gamma={\sum_{k=1}^d(n_k+m_k)r}$ is the total number of parameters. Since $f_0$ is a quadratic function of elements of $P$ and $Q$, we have $f_0 \in W_2^{\infty} [-c_0,c_0]^{\gamma}$ where $W_2^{\infty}$ is a Sobolev space, and $\|f_0\|_2=\rho(\mathbf\Theta_0,\mathbf\Theta) \le c_{10}$ for a constant $c_{10}>0$. In addition, we have $f_0^{(\alpha)}=0$ for $\alpha=\infty$. Therefore, based on Lemma 2 of \cite{shen1998method}, we get $$\|f_0\|_{\infty}=\|\mathbf\Theta_0-\mathbf\Theta\|_{\infty} \le 2c_{10}.$$ 
The required conditions are fulfilled by defining $c_9=2c_{10}$ and $\beta_2=0$.

Next, we verify the assumption on the Hellinger metric entropy. Let $$\mathcal{N}(\varepsilon,n)=\{g_1^l,g_1^u,\ldots,g_n^l,g_n^u\}$$ be a set of functions from the $L_2$ space, where $\max_{1\le i \le n} \|g_i^u-g_i^l\|_2 \le \varepsilon$. Suppose for any function $l_{\Delta} \in \mathcal{F}(k_1,k_2)$, there exists $i \in \{1,\ldots,n\}$ such that $g_i^l \le l_{\Delta} \le g_i^u$ almost surely. Then the Hellinger metric entropy is defined as $H(\varepsilon,\mathcal{F})=\log \{n: \min \mathcal{N}(\varepsilon,n)\}$.

Let $\omega=\frac{\alpha}{\gamma}=\infty$, then $p\omega=\infty>1$. Define $$\psi(k_1,k_2)=\int_{L_0}^{U_0} H^{1/2}(u,\mathcal{F})du/L_0$$ where $L_0=c_{12}\lambda_{|\Omega|}(k_1^2+k_2)$ and $U_0=c_{13}\varepsilon_{|\Omega|}(k_1^2+k_2)^{(1+\max(\beta_1,\beta_2))/2}$. Based on Theorem 5.2 of \cite{birman1967piecewise}, the Hellinger metric entropy is controlled by 
$$H(\varepsilon_{|\Omega|},\mathcal{F}) \le c_{11}\varepsilon_{|\Omega|}^{-0}=c_{11}.$$
Recall that $\beta_1=\beta_2=0$. Then for fixed $k_1$ and $k_2$, we have $\psi(k_1,k_2)=\sqrt{c_{11}} \frac{U_0-L_0}{L_0} \sim \frac{\varepsilon_{|\Omega|}-\lambda_{|\Omega|}}{\lambda_{|\Omega|}}$.
Given that $\psi \sim |\Omega|^{1/2}$, the best possible rate is achieved at $\varepsilon_{|\Omega|} \sim \lambda_{|\Omega|}^{1/2}$, that is, $$\varepsilon_{|\Omega|} \sim \frac{1}{|\Omega|^{1/2}}.$$
The result in Theorem \ref{l2convergence} then follows by applying Corollary 2 of \cite{shen1998method}.
\hfill\qedsymbol

\section*{Proof of Theorem \ref{generalconvergence}}

Let $A(k_1,k_2)$ and $\mathcal{F}(k_1,k_2)$ be defined as in the proof of Theorem \ref{l2convergence}. We start by verifying conditions of Corollary 2 of \cite{shen1998method}. First, based on the definition of $V(\cdot,\cdot)$ and Assumption \ref{smoothness}, we have 
\ba
 V(\mathbf \Theta_0,\mathbf \Theta) &=&\frac{1}{n_1\cdots n_d}\sum_{i_1=1}^{n_1}\cdots\sum_{i_d=1}^{n_d}\mbox{Var}\left\{l(\mathbf \Theta, y_{i_1\cdots i_d})-l(\mathbf \Theta_0, y_{i_1\cdots i_d})\right\}\\
&\le& \frac{1}{n_1\cdots n_d}\sum_{i_1=1}^{n_1}\cdots\sum_{i_d=1}^{n_d} \mbox{E}\{|l(\mathbf \Theta, y_{i_1\cdots i_d})-l(\mathbf \Theta_0, y_{i_1\cdots i_d})|^2\}\\
&\le& \frac{1}{n_1\cdots n_d}\sum_{i_1=1}^{n_1}\cdots\sum_{i_d=1}^{n_d} \mbox{E}\{g^2(y_{i_1 \cdots i_d})\} \|\mathbf \Theta_0-\mathbf \Theta\|^2\\
&\le& c_{2}' \|\mathbf \Theta_0-\mathbf \Theta\|^2.
\ea
Therefore, we have $\sup_{A(k_1,k_2)} V(\mathbf \Theta_0,\mathbf \Theta) \le 4c_2' k_1^2 \{1+(k_1^2+k_2)^{\beta_1}\}$ with $\beta_1=0$.

Furthermore, by Assumptions \ref{smoothness} and \ref{ball}, we have
\ba
\|l_{\Delta}(\mathbf \Theta|y_{i_1\cdots i_d})\|_2 &=& [\mbox{E}\{|l(\mathbf \Theta, y_{i_1\cdots i_d})-l(\mathbf \Theta_0, y_{i_1\cdots i_d})|^2\}]^{1/2}\\
&\le& c_2' \|\mathbf \Theta_0-\mathbf \Theta\|\\
&\le& c_{14} \rho(\mathbf \Theta_0,\mathbf \Theta)^{1+\beta}
\ea
for $c_{14}=c_2'/c_3^{1+\beta}$, a given element  $y_{i_1,\ldots,i_d}$ in the tensor, and $\delta >0$ such that $\mathbf \Theta \in B_{\delta}(\mathbf \Theta_0)$.
Then by Lemma 2 of \cite{shen1998method}, we have
\ba
\|l_{\Delta}(\mathbf \Theta|y_{i_1\cdots i_d})\|_{\infty} &\le& c_{15} \|l_{\Delta}(\mathbf \Theta|y_{i_1\cdots i_d})\|_2^{(\alpha-p^{-1})/(\alpha-p^{-1}+1/2)}\\
&\le& c_{14}c_{15}\rho(\mathbf \Theta_0,\mathbf \Theta)^{2\beta_2},
\ea
where $\beta_2=\frac{1+\beta}{2} \frac{\alpha-p^{-1}}{\alpha-p^{-1}+1/2}$. Since $\beta \in [0,1)$, we have $\beta_2 \in [0,1)$.

We now verify the condition on the Hellinger metric entropy. Recall that $\omega=\frac{\alpha}{\gamma}$. 
For any $\omega>0$, we have $p>2$ implies $\frac{p}{1-p\omega} >2$. Therefore, based on Theorem 5.2 of \cite{birman1967piecewise}, the Hellinger metric entropy is upper-bounded by
$H(\varepsilon_{|\Omega|},\mathcal{F}) \le c_{16}\varepsilon_{|\Omega|}^{-1/\omega}.$
Then we have:
\ba
\psi(k_1,k_2) &=& \int_{L_0}^{U_0} u^{-\frac{1}{2\omega}}du/L_0\\
&\sim& C(k_1,k_2) \frac{\varepsilon_{|\Omega|}^{-\frac{1}{2\omega}+1}-\lambda_{|\Omega|}^{-\frac{1}{2\omega}+1}}{\lambda_{|\Omega|}}.
\ea
The best possible rate is provided by setting $\psi(k_1,k_2) \sim |\Omega|^{1/2}$ and $\lambda_{|\Omega|} \sim \varepsilon_{|\Omega|}^2$. Hence,
\ba
\varepsilon_{|\Omega|}^{-\frac{1}{2\omega}-1}-\varepsilon_{|\Omega|}^{-\frac{1}{\omega}} \sim |\Omega|^{1/2}.
\ea
That is,
\[ \varepsilon_{|\Omega|} \sim
  \begin{cases}
(\frac{1}{{|\Omega|}^{1/2}})^{\frac{2\omega}{2\omega+1}}       & \mbox{if } \omega>\frac{1}{2}\\
(\frac{1}{{|\Omega|}^{1/2}})^{\omega}  & \mbox{if } \omega\leq\frac{1}{2}
  \end{cases}.
\]
Then the result follows by applying Corollary 2 of \cite{shen1998method}.
\hfill\qedsymbol

\bibliographystyle{acmtrans-ims}
\bibliography{tensor}

\end{document}